\documentclass{article}

\usepackage[utf8]{inputenc}
\usepackage[T1]{fontenc}
\usepackage[margin=1in]{geometry}
\usepackage{amsmath, amssymb, bm}
\usepackage{xcolor}
\usepackage{url}

\usepackage{amsmath,amsfonts,bm}

\def\eqref#1{equation~\ref{#1}}

\def\1{\bm{1}}

\DeclareMathAlphabet{\mathsfit}{\encodingdefault}{\sfdefault}{m}{sl}
\SetMathAlphabet{\mathsfit}{bold}{\encodingdefault}{\sfdefault}{bx}{n}

\usepackage{algorithm}
\usepackage{algpseudocode}
\usepackage{booktabs}
\usepackage{graphicx}
\usepackage{subcaption}
\usepackage{multirow} 
\usepackage{siunitx}

\graphicspath{ {Figures/} }

\usepackage{listings}

\definecolor{codebg}{RGB}{245,245,245}
\definecolor{framegray}{gray}{0.85}
\definecolor{keycolor}{RGB}{17,122,101}     
\definecolor{listval}{RGB}{150,150,150}     

\lstdefinelanguage{jsonplus}{
  morestring=[b]",
  sensitive=true,
  alsoletter={\_,:},
  morecomment=[l]{//},
  morekeywords={true,false,null},
}

\lstdefinestyle{promptstyle}{
  backgroundcolor=\color{codebg},
  basicstyle=\ttfamily\scriptsize,
  frame=single,
  rulecolor=\color{framegray},
  frameround=tttt,
  xleftmargin=1em,
  xrightmargin=1em,
  language=jsonplus,
  columns=fullflexible,
  moredelim=**[is][\color{keycolor}]{@k}{@},
  moredelim=**[is][\color{listval}]{@v}{@},
}

\usepackage[accsupp]{axessibility}  

\usepackage[pagebackref,breaklinks,colorlinks]{hyperref}

\title{Discovering Intersectional Bias via Directional Alignment in \\ Face Recognition Embeddings}

\author{Ignacio Serna \\
Center for Humans \& Machines \\
Max Planck Institute for Human Development, Berlin, Germany \\
\texttt{serna@mpib-berlin.mpg.de}
}

\providecommand{\keywords}[1]
{	
  \textbf{\textit{Keywords---}} #1
}

\begin{document}

\maketitle

\begin{abstract}
  Modern face recognition models embed identities on a unit hypersphere, where identity variation forms tight clusters. Conversely, shared semantic attributes can often be effectively approximated as linear directions in the latent space. Existing bias evaluation methods rely on predefined attribute labels, synthetic counterfactuals, or proximity-based clustering, all of which fail to capture intersectional subpopulations that emerge along latent directions. We introduce LatentAlign, an attribute-free algorithm that discovers semantically coherent and interpretable subpopulations by iteratively aligning embeddings along dominant latent directions. Unlike distance-based clustering, LatentAlign exploits the geometry of hyperspherical embeddings to isolate directional structures shared across identities, allowing for the interpretable discovery of attributes. Across four state-of-the-art recognition backbones (ArcFace, CosFace, ElasticFace, PartialFC) and two benchmarks (RFW, CelebA), LatentAlign consistently yields more semantically coherent groups than $k$-means, spherical $k$-means, nearest-neighbor search, and DBSCAN. Crucially, the discovered subpopulations expose severe intersectional vulnerabilities, with False Match Rates up to 4× higher than groups defined by explicit annotations. Our results show that by treating semantic attributes as directional features rather than spatial clusters, we can effectively isolate intersectional subpopulations and expose hidden biases that standard audits miss.
\end{abstract}

\keywords{Face recognition, Bias, Fairness, Interpretability, Safety}

\section{Introduction}

Face recognition technology (FRT) has seen widespread adoption in applications ranging from unlocking smartphones and securing sensitive facilities to assisting law enforcement and aiding in search operations. Rapid advancements in FRT have driven accuracy to levels that exceed human performance \cite{deng2019arcface,liu2017sphereface,phillips2018face,schroff2015facenet}, promising enhanced efficiency and error reduction. Yet, as these systems proliferate, so do the risks associated with their misuse, ranging from privacy breaches to civil rights violations \cite{castelvecchi2020facial,van2020ethical,civilrights2024FR}. Numerous studies have highlighted systematic biases in these systems \cite{cavazos2020accuracy,drozdowski2020bias,garcia2019harms,wang2020mitigating,gong2021gac,robinson2020facebias,serna2022sensitive,terhorst2021comprehensive,yucer2024racial}. These biases can result in unfair treatment, exacerbating societal inequalities and undermining trust in their use \cite{garvie2016perpetual,clayton2024bbc}.

Bias in face recognition algorithms refers to systematic differences in performance that affect certain groups of people based on characteristics such as race, gender, age, or other demographic factors. These biases creep in at various stages of the algorithm's lifecycle, from data collection and preprocessing to model design.

Traditionally, bias identification and mitigation rely on supervised approaches, where labeled datasets are used to define and evaluate group-specific performance gaps. However, such methods depend on explicit annotations \cite{terhorst2021comprehensive}, which may be unavailable, costly, or prone to unintended biases introduced by human annotators and unnoticed confounders \cite{balakrishnan2021towards}. This approach faces a "label bottleneck": it is limited to fixed taxonomies, and blinded to intersectional biases (e.g., specific combinations of age, phenotype, and accessories) that do not fit predefined categories. While synthetic datasets have been proposed as an alternative \cite{perona2023benchmarking}, they still only account for predefined attributes, leaving room for unconsidered biases to persist.

To circumvent the reliance on labels, practitioners have turned to unsupervised clustering to discover error-prone subgroups \cite{sohoni2020no, krishnakumar2021udis}. We argue, however, that these methods suffer from a fundamental geometric mismatch. Modern face recognition models constrain feature embeddings to lie on a unit hypersphere, and are trained with angular margin losses to optimize the clustering of identities. This effectively fragments the latent space into millions of tight, identity-specific clusters. Standard clustering algorithms fail here because they inherently converge on these dense identity regions, whereas semantic variations manifest as feature directions \cite{vitomir2024discovering} rather than distinct clusters. This results in either identity groupings or semantically incoherent noise.


The challenge of identifying and analyzing bias in face recognition models has thus been largely constrained to known, predefined attributes. We address this limitation by introducing LatentAlign, an unsupervised algorithm designed explicitly for the geometry of deep face embeddings. Instead of clustering based on proximity, LatentAlign identifies subpopulations by iteratively aligning samples along dominant latent directions, allowing us to isolate highly specific, intersectional subpopulations that standard clustering misses. This method autonomously uncovers subpopulations correlated with performance disparities, without requiring attribute annotations or synthetic manipulations.

\begin{figure}[t]
    \centering
    \begin{subfigure}{0.34\linewidth}
        \centering
        \includegraphics[width=\linewidth]{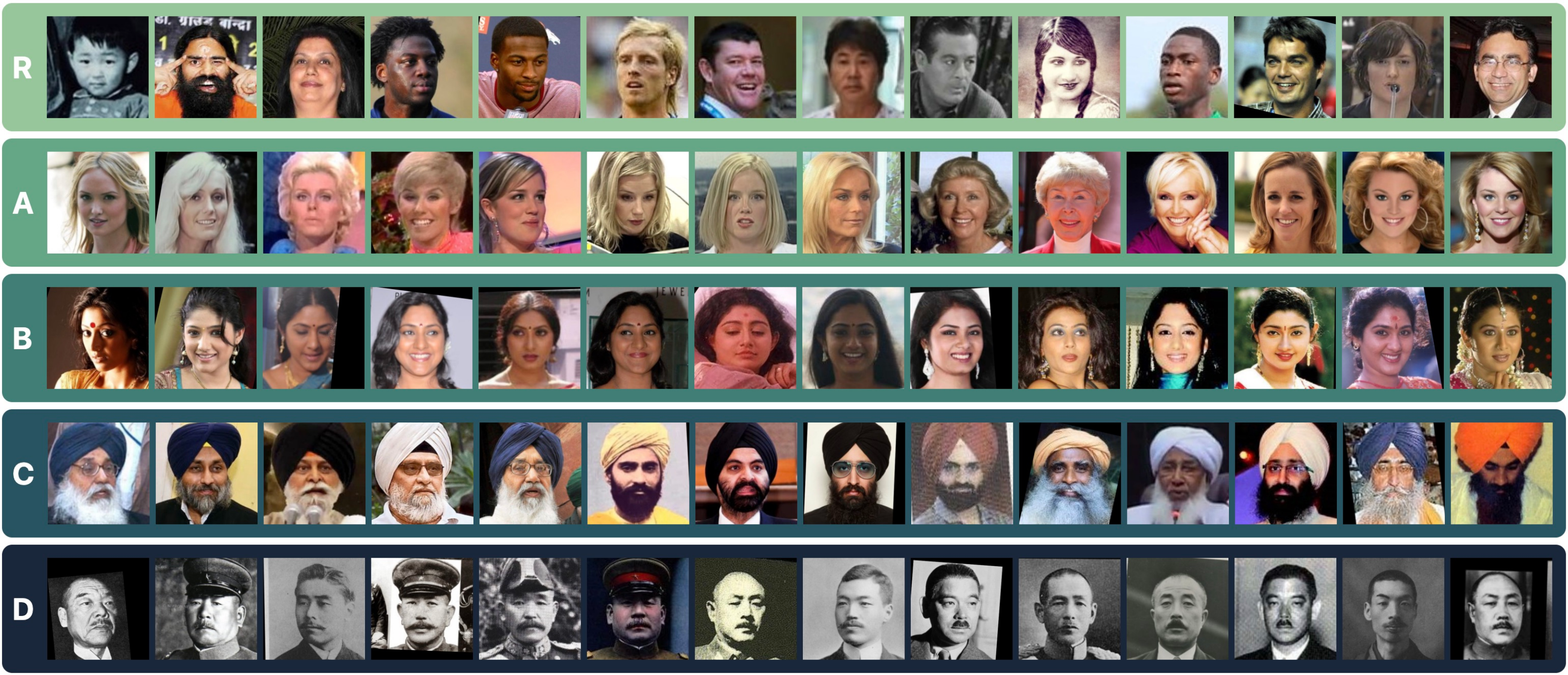}
        \caption{}
        \label{fig:example-images}
    \end{subfigure}
    \begin{subfigure}{0.32\linewidth}
        \centering
        \includegraphics[width=\linewidth]{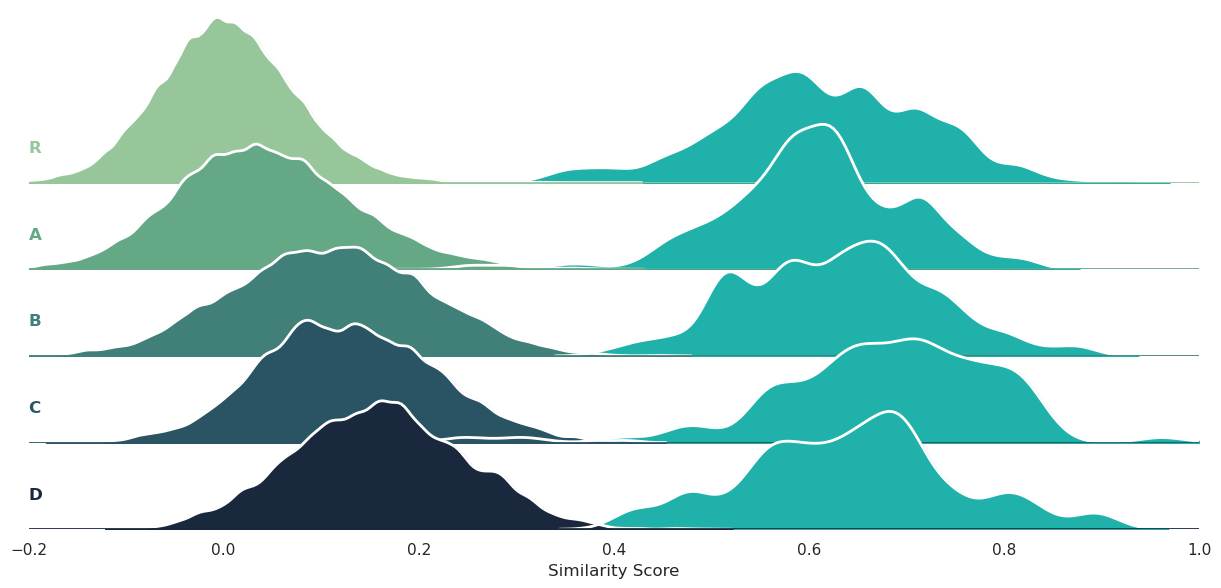}
        \caption{}
        \label{fig:distributions}
    \end{subfigure}
    \begin{subfigure}{0.32\linewidth}
        \centering
        \includegraphics[width=\linewidth]{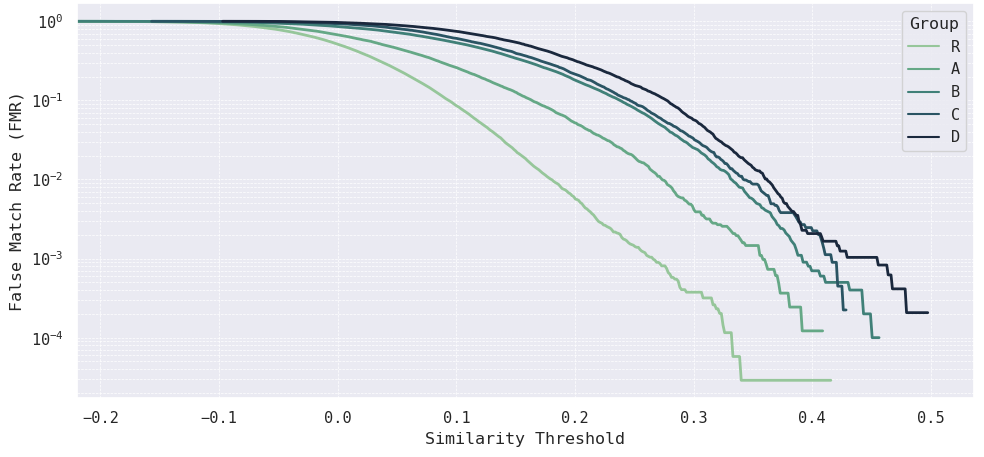}
        \caption{}
        \label{fig:FMRcurve}
    \end{subfigure}
    \caption{\textbf{Intersectional biases discovered by LatentAlign on ArcFace.}
            \textbf{Left:} Example images from different groups (R: random sample, A: Caucasian blonde females, B: young Indian women, C: Indian males with turbans, D: Chinese men with mustaches in black-and-white photographs). 
            \textbf{Center:} Similarity score distributions for groups R–D. The right distribution of each group corresponds to genuine pairs (higher similarity), while the left one represents impostor pairs (lower similarity). Darker colors indicate higher error rates. 
            \textbf{Right:} False Match Rate (FMR) curves for groups R–D, computed from genuine and impostor distributions (center). The extent of bias is indicated by the hue of the color used; darker hues denote a higher degree of bias.}
\end{figure}

Figure~\ref{fig:example-images} shows images of different groups found with LatentAlign on ArcFace embeddings. Figure \ref{fig:distributions} represents the distributions of cosine-similarity scores for genuine (right distribution) and imposter (left distribution) pairs within each group. A genuine score is computed with a pair of embeddings belonging to the same identity and an imposter score with embeddings from different identities. Figure \ref{fig:FMRcurve} contains the FMR (False Match Rate) curve for each group. These curves represent the probability that the system incorrectly mismatches two different individuals as a function of the similarity threshold. The gap between the FMR curves is the sign of bias across groups. Other models have similar score distributions and similar error rate curves (see Appendix \ref{Appendix: Evaluation} in the supplementary material for detailed plots).


Using latent directions provides two key benefits over standard clustering: (i) \textbf{semantically coherent grouping}, where samples sharing common attributes are grouped more reliably than by distance-based methods, and (ii) \textbf{discovery of interpretable directions}, which align with semantic attributes such as age, ethnicity, or attire and expose bias-related subpopulations.

Our contributions are:
\begin{itemize}
    \item \textbf{Geometric Alignment Algorithm:} We propose LatentAlign, a method that discovers semantically coherent subpopulations in face recognition embeddings by aligning samples along latent directions. We validate that the directions discovered by LatentAlign are semantically meaningful using generative traversals (Arc2Face, InfiniteYou).

    \item \textbf{Superior Semantic Coherence:} We demonstrate that LatentAlign consistently yields groups with lower intra-group attribute distance compared to $k$-means, spherical $k$-means, DBSCAN, and nearest-neighbor search across four major recognition backbones (ArcFace, CosFace, ElasticFace, PartialFC) and two databases (RFW and CelebA).

    \item \textbf{Bias Discovery:} We show that the subpopulations found by LatentAlign exhibit False Match Rates (FMR) up to 4$\times$ higher than explicit annotations baselines, exposing biases masked by broad demographic labels.
\end{itemize}

\section{Related Work}

Our work intersects two central areas of research: (1) the discovery and interpretation of latent representations in computer vision models, and (2) the detection of bias in deep learning models. Prior work in these domains can be broadly classified by their supervision paradigm (supervised vs. unsupervised) and their application focus (interpretability or bias mitigation).

\subsection{Supervised Approaches}

\textbf{Latent Concept Discovery in Vision Models.}  
Early efforts to interpret neural networks revealed that human-defined concepts can often be represented as directions in latent space rather than as isolated neurons. Notably, \textit{Concept Activation Vectors} (CAVs) \cite{kim2018interpretability} are learned by training a linear classifier to distinguish between concept-specific activations and random activations. Subsequent work has extended this idea to semi-supervised settings, using latent space decomposition to uncover concept vectors \cite{graziani2023uncovering,graziani2023disentangling}.

\textbf{Interpretable Feature Directions in Face Recognition.}  
\cite{vitomir2024discovering} explored interpretability in face recognition by identifying semantically meaningful directions in embedding spaces. Their method relied on annotated facial attributes and blended-image techniques to recover latent directions, which were then traversed for qualitative interpretation.

\textbf{Synthetic Image Generation for Bias Detection.}  
Counterfactual data augmentation has been proposed to diagnose bias by generating synthetic faces that selectively vary target attributes \cite{balakrishnan2021towards,hutchinson2019detecting,joo2020gender,perona2023benchmarking}. While these methods support causal analyses, they are restricted to attributes chosen a priori, limiting their ability to uncover unexpected sources of bias.

\subsection{Unsupervised Approaches}

\textbf{Discovery of Latent Semantics in Generative Models.} 
Advances in Generative Adversarial Networks (GANs) have significantly enhanced our ability to interpret and manipulate latent spaces, leading to controlled and meaningful image generation \cite{harkonen2020ganspace}. For example, SeFa \cite{bolei2021semantics} introduced a closed-form factorization for uncovering latent semantic directions without additional training or sampling. Other approaches enforce orthogonality and distinctiveness of latent directions \cite{song2023householder} or use contrastive objectives \cite{yuksel2021latentclr}.  And others learn clusters corresponding to meaningful attributes
\cite{georgopoulos2022cluster}. These approaches demonstrate the promise of latent direction analysis, though they are generally applied to generative rather than discriminative models. 

\textbf{Unsupervised Bias Discovery via Clustering.}  
Clustering-based methods have been proposed to detect bias in the absence of labels. For example, \cite{krishnakumar2021udis} used hierarchical clustering of embeddings to identify subgroups with degraded classification performance in models trained for image classification. \cite{gluge2020not} found simple clustering measures unreliable for detecting face recognition bias, arguing for more nuanced approaches.

These findings highlight both the promise and the limitations of unsupervised grouping for bias auditing: while clustering can reveal potential subgroups of concern, naïve distance-based methods may fail to align with semantically meaningful attributes or with actual performance disparities. Our method addresses these challenges by introducing \textit{LatentAlign}, which forms groups along latent directions in the embedding space. This provides higher semantic coherence than local similarity-based clustering, while simultaneously uncovering interpretable directions that correspond to meaningful, previously unlabeled factors of variation.

\section{Methodology}

\subsection{LatentAlign: Iterative Semantic Discovery}

At the core of our approach lies LatentAlign, an algorithm that discovers semantically coherent subpopulations of face images by iteratively estimating and aligning latent directions in the embedding space. Instead of clustering points based on proximity, LatentAlign operates as a self-correcting discovery process, where it iteratively refines a latent direction vector.

Let $\mathcal{X}=\{x_i\}_{i=1}^N$ be a dataset of $N$ face images mapped to $\ell_2$-normalized embeddings $\vec{\bm{e}}_i \in \mathbb{R}^{d}$ via a deep feature extractor $\phi(\cdot)$. Thus, the embedding function can be seen as a mapping
\[
\phi : \mathcal{X} \to \mathbb{S}^{d-1},
\]
where $\mathbb{S}^{d-1}$ denotes the unit hypersphere in $\mathbb{R}^{d}$. Let $\ell_i$ denote the identity label of $x_i$. Given an initial seed group of indices $\mathcal{S}$, the algorithm first computes a latent direction $\vec{\bm{v}}$ that represents the group's shared attributes.

\textbf{Drift Prevention via Inverse-Frequency Weighting.} Face recognition datasets naturally contain highly unbalanced numbers of images per identity. If we were to use a naive average to find the group's direction, the vector would "drift" toward facial features of the most dominant identity. To mitigate this, LatentAlign introduces a drift-prevention mechanism: an inverse-frequency weighting scheme that equalizes the contribution of each identity. The robust latent direction is computed as:

\begin{equation}
    \tilde{\vec{\bm{v}}} = \frac{1}{C} \sum_{j\in \mathcal{S}} w_j \vec{\bm{e}}_j
    \;, \qquad
    \vec{\bm{v}} = \frac{\tilde{\vec{\bm{v}}}}{\|\tilde{\vec{\bm{v}}}\|_2}
\end{equation}

and then $\ell_2$-normalized to lie on the unit hypersphere. Where $w_j = \frac{1}{c_{\ell_j}}$ is the weight of sample $j$, $c_{\ell_j}$ is the count of samples with identity ${\ell_j}$ in $\mathcal{S}$, and $C$ is the number of unique identities in $\mathcal{S}$. 

\textbf{Iterative Alignment and Refinement.} Once the initial direction $\vec{\bm{v}}$ is established, the algorithm searches the broader dataset for the embedding most aligned with this vector. Alignment is quantified by projecting candidate embeddings onto $\vec{\bm{v}}$:

\begin{equation}
    \textit{i*}= \underset{k \notin \mathcal{S}}{\operatorname{argmax}} \; \langle\vec{\bm{e}}_k , \vec{\bm{v}}\rangle
\end{equation}

The most aligned sample $\textit{i*}$ is added to the group $\mathcal{S} = \mathcal{S} \cup \{ \textit{i*}\}$, and the direction $\vec{\bm{v}}$ is immediately recomputed. 

Since both embeddings ($\vec{\bm{e}}_k$) and current latent direction ($\vec{\bm{v}}$) are $\ell_2$-normalized, projection equals cosine similarity. Note that by projecting the embeddings onto the latent vector, it captures the extent to which that direction is represented in each image. Since all embeddings are normalized, the embedding with the largest projection corresponds to the one with the greatest component of $\vec{\bm{v}}$ in its direction.

\begin{algorithm}
    \caption{\textsc{Latent Alignment}}
    \label{alg:LFA}
    \begin{algorithmic}[1]
        \Require Dataset embeddings $\bm{E}$, initial group $\bm{S}$
        \While{$\max\{\bm{p}\} \geq \tau$}
            \State $\vec{\bm{v}} \gets$ \textsc{GetLatentDirection}($\bm{S}$)
            \State $\bm{p} \gets \bm{E} \cdot \vec{\bm{v}} \;$ \Comment{Project embeddings onto direction}
            \State $i \gets \arg\max \{\bm{p}\}$ \Comment{Most aligned embedding}
            \State $\bm{S} \gets \bm{S} \cup \{\vec{\bm{e}}_i\}$ \Comment{Add to group}
            \State $\bm{E} \gets \bm{E} \setminus \{\vec{\bm{e}}_i\}$ \Comment{Remove}
        \EndWhile
        \State \Return $\bm{S}$
    \end{algorithmic}
\end{algorithm}

\begin{algorithm}
    \caption{\textsc{GetLatentDirection}}
    \label{alg:GetLatentDirection}
    \begin{algorithmic}[1]
        \Require Embeddings $\{\vec{\bm{e}}_j\}_{j=1}^n$ and identity labels $\{\,\ell_j\}_{j=1}^n$
        \Ensure Latent direction $\vec{\bm{v}}$
        \State $\mathcal I \gets \{\ell_j : j=1,\dots,n\}$ \Comment{Set of unique identities}
        \For{each identity $i \in \mathcal I$}
            \State $c_i \gets |\{\,j : \ell_j = i\}|$ \Comment{Count samples per identity}
        \EndFor
        \For{$j=1,\dots,n$}
            \State $w_j \gets 1 / c_{\ell_j}$ \Comment{Inverse frequency weighting}
        \EndFor
        \State $\tilde{\vec{\bm{v}}} \gets \sum_{j=1}^n w_j\,\vec{\bm{e}}_j$ \Comment{Weighted sum}
        \State $\vec{\bm{v}} \gets \tilde{\vec{\bm{v}}} \;/\; \|\tilde{\vec{\bm{v}}}\|_2$ \Comment{Normalize}
        \State \Return $\vec{\bm{v}}$
    \end{algorithmic}
\end{algorithm}

This iterative loop is the engine of the algorithm's semantic coherence. By continuously recalculating $\vec{\bm{v}}$ as new, highly aligned samples are added, the vector is dynamically refined. As the subpopulation grows, identity-specific variation is progressively attenuated, and the latent direction increasingly reflects attributes shared across the discovered group. This self-correcting process repeats until the maximum projection score falls below a predefined threshold ($\tau$). The threshold $\tau$ controls the semantic tightness of the discovered group and is set consistently across experiments (see Section \ref{Appendix: Ablation Study} for an extended analysis).

\textbf{Initialization.} LatentAlign requires an initial seed group $\mathcal{S}$. While exhaustive enumeration of small subsets could mean finding all semantic directions, it is computationally intractable for datasets like CelebA or RFW. Therefore, we utilize a simple heuristic: we extract highly connected components from a sparse cosine-similarity graph (threshold = 0.5) to serve as initial seeds.
Please refer to the Appendix \ref{Appendix: Ablation Study} for a complete ablation study on the heuristic threshold.

\subsection{Application: Unsupervised Bias Auditing}

While LatentAlign is a general tool for discovering semantic directions, our primary application in this work is auditing face recognition models for hidden biases. Once LatentAlign has autonomously grown a seed into a fully formed subpopulation, we evaluate it for systematic performance disparities.

Face recognition models are trained such that embeddings of the same identity are closely clustered, while different identities are mapped far apart. However, in biased subpopulations, distinct identities are closer to each other. This degradation can be quantified using impostor similarity scores (the cosine similarity between embeddings belonging to different identities). Elevated average impostor similarity within a group indicates a lack of discriminative power, leading to increased false match rates. Within this framework, LatentAlign acts as the bridge between coherent group discovery and the systematic identification of their failure.

\section{Experimental Evaluation} \label{sec:evaluation}

We assess the effectiveness of our proposed method, LatentAlign, in identifying semantically meaningful and potentially biased subpopulations. Our evaluation focuses on three key aspects: (1) the interpretability of latent feature directions, (2) the semantic coherence of the discovered groups, and (3) the relationship between the discovered groups and performance bias.

\subsection{Datasets and Models}

We conduct our experiments on the following datasets:

\textbf{RFW} \cite{wang2019racial}: A benchmark designed to measure racial bias in face recognition systems. It contains four subsets—Caucasian, African, Asian, and Indian—each with approximately 3,000 identities and 10,000 images. Its primarily source was the MS-Celeb-1M \cite{2016msceleb} dataset.

\textbf{CelebA} \cite{liu2015faceattributes}: A large-scale dataset consisting of over 200,000 images of more than 10,000 celebrities, annotated with 40 binary attributes. It serves as a benchmark for attribute-based classification.

For our face embeddings, we use four widely adopted face recognition models: ArcFace \cite{deng2019arcface}, CosFace \cite{wang2018cosface}, Partial FC \cite{an2021partial}, and ElasticFace \cite{boutros2022elasticface}.

We annotate the RFW dataset using state-of-the-art open-source Vision-Language Models (VLMs), including InternVL3 \cite{chen2024internvl}, Ola \cite{liu2025ola}, Ovis2 \cite{lu2024ovis}, Qwen2.5-VL \cite{Qwen2.5-VL}, and SAIL-VL \cite{dong2025scalable}. Final labels are obtained via majority voting. To assess reliability, we manually annotate a 1K-image subset, obtaining substantial to almost perfect agreement between human and VLM annotations (Cohen’s $\kappa$ values between 0.75 and 0.97 across attributes). The resulting labels cover 10 demographic and appearance attributes. Comprehensive details regarding the annotation and validation procedures are available in Appendix \ref{Appendix: Labels}.

\begin{figure}[t]
    \centering
    \begin{subfigure}{0.59\linewidth}
        \centering
        \includegraphics[width=\linewidth]{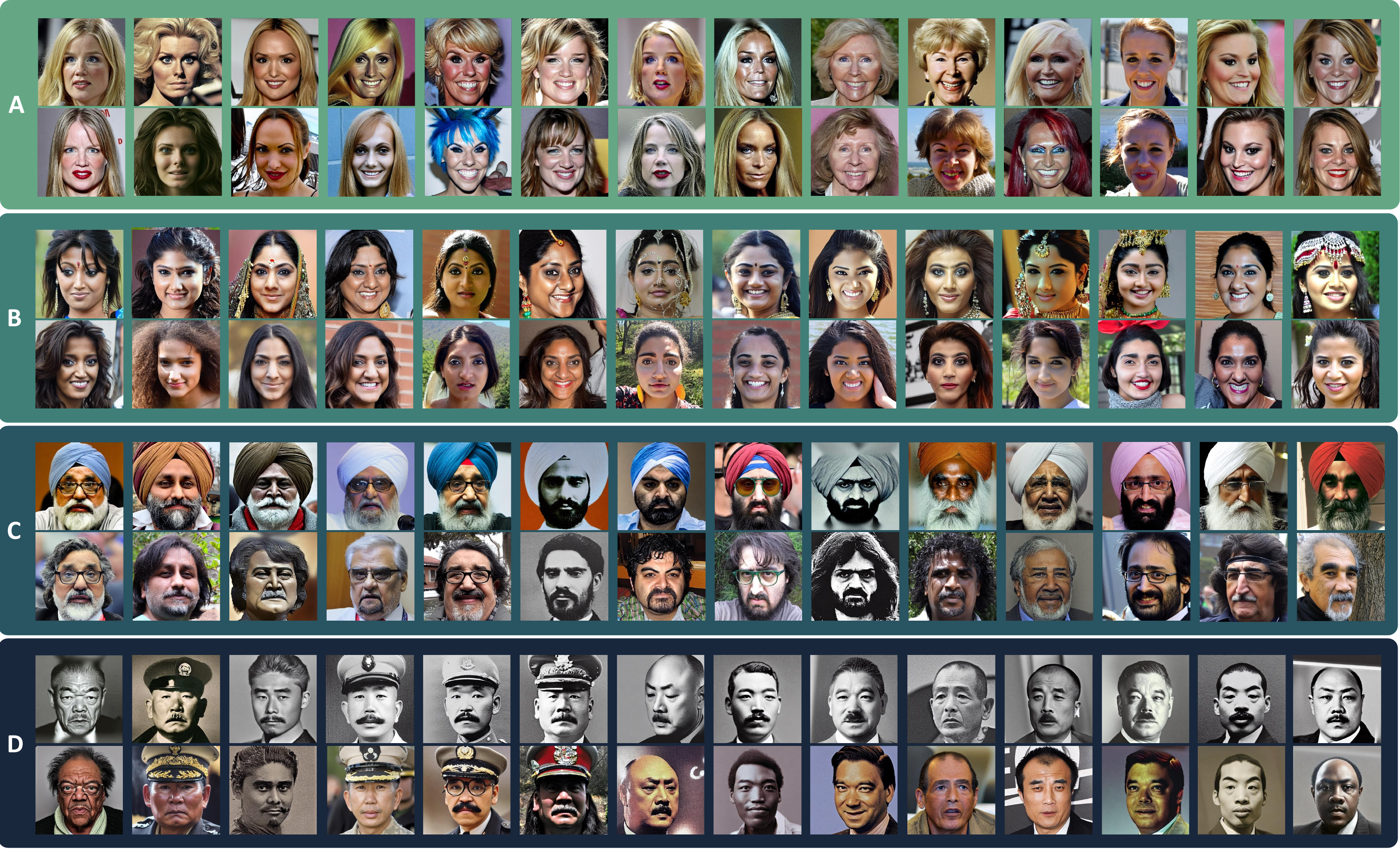}
        \caption{Removal}
        \label{fig:subtraction}
    \end{subfigure}\hfill
    \begin{subfigure}{0.4\linewidth}
        \centering
        \includegraphics[width=\linewidth]{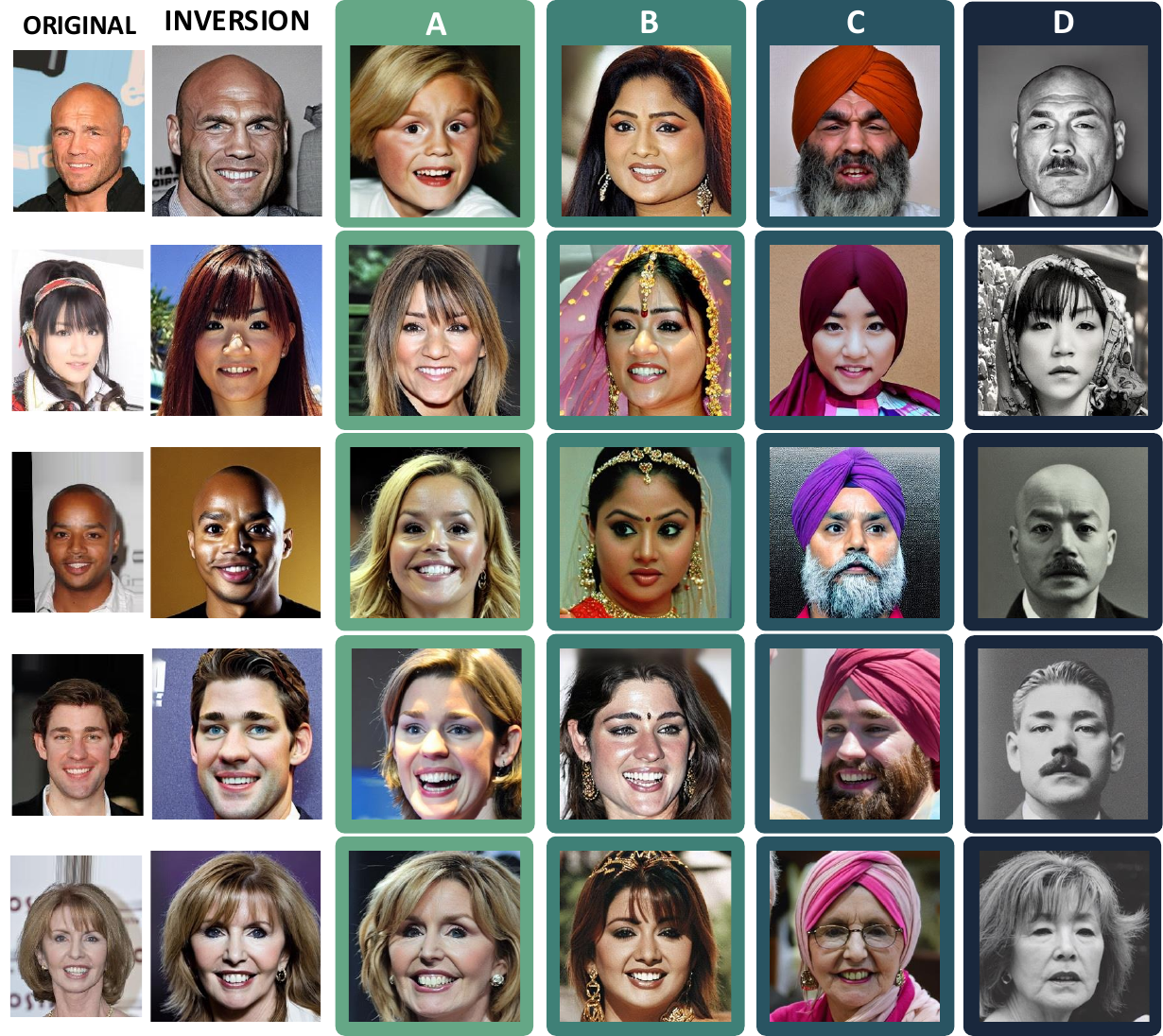}
        \caption{Addition}
        \label{fig:addition}
    \end{subfigure}
    \caption{Arc2Face \cite{papantoniou2024arc2face} generations. \textbf{Left: Removal of semantic attributes from face images}. For each group (A, B, C and D) the top image shows the inversion (embedding decoded) of the original image and the bottom one the original image after traversing back the embedding along the latent direction of the group. \textbf{Right: Addition of semantic attributes}. From left: first column shows the original image, second column the inversion, and the rest of the columns (A-D) the original image decoded after traversing it along the latent direction of groups A-D.}
\end{figure}

\subsection{Validating the Geometry: Interpretable Latent Directions}
\label{Sec: Interpretability}

We first validate our core hypothesis that these latent vectors encode coherent semantics. To this end, we traverse each latent direction in the embedding space and decode the resulting embeddings into images, allowing us to visually inspect how identity and attributes change along each direction.

To ensure that our interpretability analyses are not biased by specific architectural choices, we deliberately employ 
two identity-preserving generative models with fundamentally different designs: Arc2Face \cite{papantoniou2024arc2face}, which uses ArcFace-R100 and FLUX.1; and InfiniteYou \cite{jiang2025infiniteyou}, that uses IR-SE50 and Stable Diffusion. Both synthesize high-quality face images conditioned on an identity embedding.

We evaluate the semantic validity of the discovered directions by selecting groups formed on the RFW database and traversing their latent directions. Moving along the negative direction removes attributes from the group's face images, while moving along the positive direction synthesizes those attributes onto different face images. Details on the methodology for traversing the embedding space can be found in Appendix \ref{Appendix: Traversals}.

The qualitative visual results are shown in Figures \ref{fig:subtraction} and \ref{fig:addition}. Figure \ref{fig:subtraction} shows the four groups with their images decoded with Arc2Face, with the top row of each group showing the original embedding decoded (called inversion), and the bottom row decoded after being traversed along the negative direction. Figure \ref{fig:addition} shows random images from CelebA dataset (left-most column), their inversion (second left-most column), and the decoded image after traversing it along the positive direction of groups A, B, C and D. Note that these directions were discovered using images from one database (RFW) but applied to images from another database (CelebA), this demonstrates that directions are a property of the latent space produced by the model and not tied to a particular database.

They reveal that latent directions discovered by LatentAlign correspond to intersectional groups: group A is composed of white blonde females, group B are young Indian women, group C senior Indian males with turbans, and group D Chinese men with mustaches in black-and-white.

To quantify the extent to which these directions correspond to meaningful attributes we train an attribute classifier on CelebA following \cite{lingenfelter2021improving} and measure facial-attribute probabilities as embeddings are traversed with different step sizes (interpolation strengths).

\begin{figure}[t]
    \begin{subfigure}{0.49\linewidth}
        \centering
        \includegraphics[width=\linewidth]{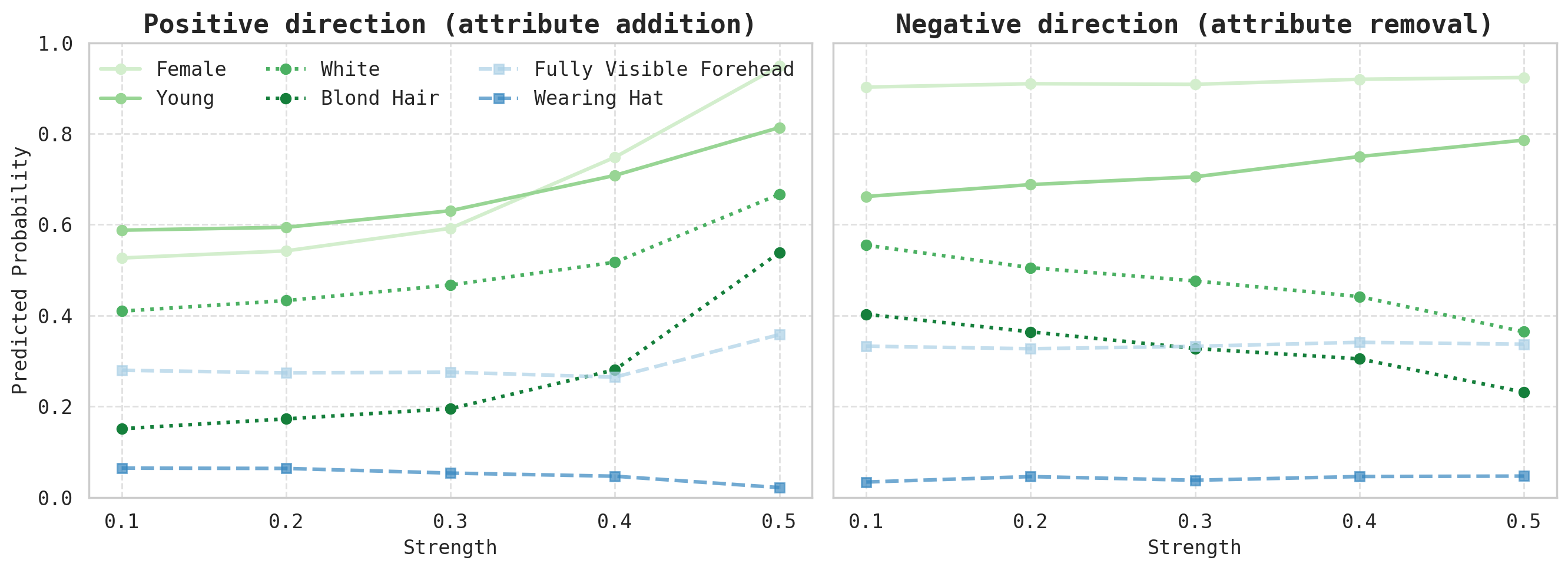}
        \caption{Arc2Face}
    \end{subfigure}
    \begin{subfigure}{0.49\linewidth}
        \centering
        \includegraphics[width=\linewidth]{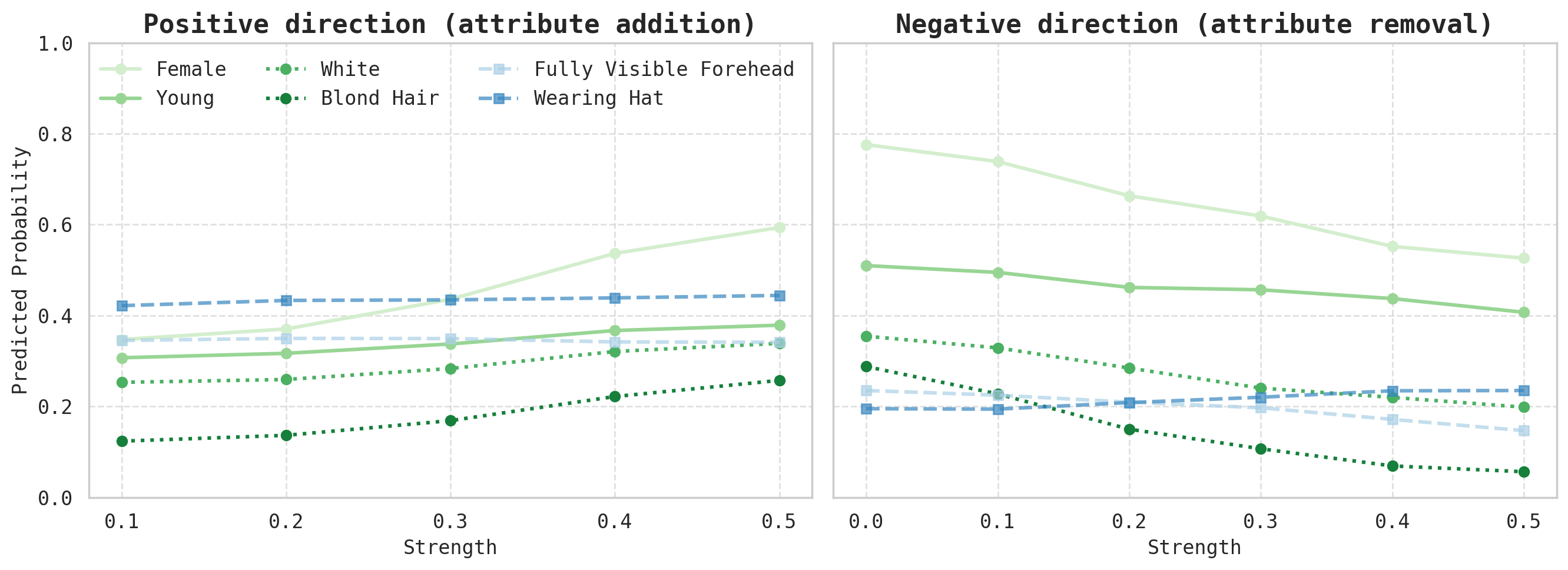}
        \caption{InfiniteYou}
    \end{subfigure}
    \caption{\textbf{Attribute probability vs traversal strength}. Predicted probabilities from an attribute classifier for images traversed along the latent direction of Group A, and encoded-decoded with Arc2Face (a) or InfiniteYou (b). Relevant attributes (e.g., Blond Hair, Young, Female, White) change monotonically with traversal strength, while unrelated attributes (e.g., Wearing Hat, Fully Visible Forehead) remain stable.}
    \label{fig:quantitative}
\end{figure}

Figure \ref{fig:quantitative} reports the predicted probabilities of several facial attributes of images traversed along the latent direction identified by LatentAlign on Group A of RFW. Arc2Face (a) and InfiniteYou (b) generative models are used. Traversals in the positive direction (attribute addition) were performed on 1,000 random CelebA samples, and in the negative (attribute removal) were performed on the Group A of RFW. The plots reveal meaningful and monotonic changes in relevant attributes (e.g., Blond Hair, Young, Female, White) while unrelated attributes (e.g., Wearing Hat, Fully Visible Forehead) remain stable.

While the overarching semantic directions remain highly consistent across both models, we observe distinct decoder-specific traversal asymmetries. For instance, Arc2Face exhibits unique anomalies such as \textit{Young} increasing in both positive and negative traversals, and \textit{Female} remaining flat in the negative direction despite increasing in the positive one—patterns not strictly replicated by InfiniteYou. Similarly, moving in the direction of a female group (Groups A and B) alters male gender, but moving in the direction of a male group (Groups C and D) does not change female gender. Furthermore, traversal effectiveness appears sensitive to identity and image quality, as seen in the inconsistent mustache removal in Group D.

These deviations suggest that while LatentAlign isolates valid semantic directions, individual decoders introduce unique modeling constraints—likely stemming from non-linear embedding structures or uneven decoder coverage. Nevertheless, despite these localized inconsistencies, both InfiniteYou and Arc2Face preserve highly consistent global direction semantics. 

Note that traversing any direction in an arbitrary manner results in the decoding of a meaningless or generic image that has nothing to do with the original identity. The results are surprising since our algorithm operates on discriminative face recognition embeddings, which are optimized for verification accuracy and not trained for semantic editing like generative models.

\subsection{Semantic Coherence: The Geometric Advantage}
\label{Sec: Semantic Coherence}

\begin{figure}[t]
    \centering
    \includegraphics[width=\linewidth]{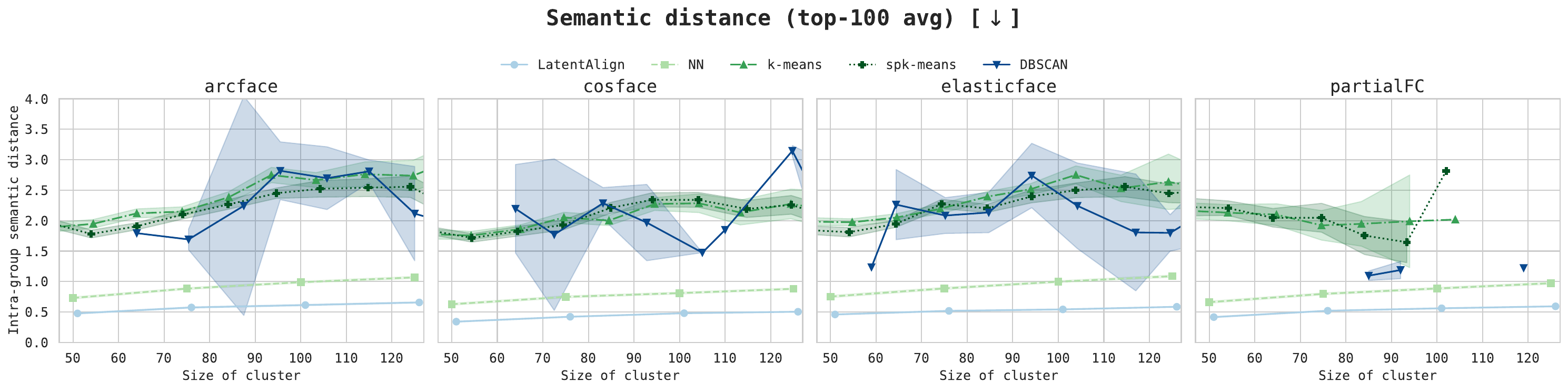}
    \caption{\textbf{Semantic coherence of groups discovered by each algorithm on RFW.} For each algorithm and FR backbone, we aggregate all discovered groups from all parameter settings, bin them by group size, and compute the average intra-group semantic distance of the top-100 most coherent groups (lower = more semantically consistent). LatentAlign consistently identifies groups with stronger semantic coherence.}
    \label{fig:semantic_comparison}
\end{figure}

Having established that latent directions encode explicit, interpretable semantics, we now evaluate whether LatentAlign's directional approach forms superior subpopulations compared to standard proximity-based methods.

\subsubsection{Baselines and Metrics.} We benchmark LatentAlign against traditional clustering algorithms, applied to the face embeddings: $k$-means, DBSCAN, nearest-neighbor (NN) search, and crucially, spherical $k$-means. We evaluate performance using Semantic Distance, defined as the average pairwise
Hamming distance of attributes within a discovered group. A lower coherence distance indicates a tighter, more semantically uniform subpopulation.

To ensure fair comparison across algorithms with fundamentally different behaviors, we control for group size. In Table \ref{tab:coherence}, we configure all methods to yield an average group size of $N \sim 100$. LatentAlign nearly halves the semantic distance on RFW compared to the baselines. Distances are higher overall on CelebA due to its larger 40-binary-attribute taxonomy and known label noise \cite{hand2018doing,lingenfelter2022quantitative,wu2023consistency}, compared to the 10 classes of the RFW annotations, yet LatentAlign maintains a clear advantage.

To confirm this performance is not an artifact of specific parameter choices, Figure \ref{fig:semantic_comparison} presents a rigorous parameter sweep on RFW. We aggregated all discovered groups across all parameter combinations, binned them by size, and evaluated the top-100 most coherent groups per bin. Across all evaluated group sizes and model backbones, LatentAlign consistently discovers the most cohesive and semantically similar groups (lower intra-group attribute distance). 

We note that optimizing $k, \epsilon$ with standard clustering indices (e.g., Silhouette, Calinski-Harabasz) recovers the number of identities, which is unsuitable for our goal of discovering semantic attributes rather than identity partitions. Spherical $k$-means serves as our strongest baseline because it also operates on cosine similarity, allowing us to isolate the benefit of LatentAlign's iterative directional alignment versus static centroid clustering. 

\begin{table}[t]
    \centering
    \caption{Semantic distance (↓ average attribute distance) across CelebA and RFW datasets. 
    LatentAlign thresholds and $k$ values were chosen to yield comparable group sizes ($N \approx 100$). 
    Across both datasets, LatentAlign is expected to produce more semantically coherent groups than nearest-neighbor search (NNS), $k$-means, and spherical $k$-means (sp$k$-means).}
    \label{tab:coherence}
    \resizebox{\columnwidth}{!}{%
        \setlength{\tabcolsep}{3pt}
        \begin{tabular}{lcccccccc}
             & \multicolumn{4}{c}{CelebA} & \multicolumn{4}{c}{RFW} \\
            \cmidrule(lr){2-5} \cmidrule(lr){6-9}
            Method &
            \footnotesize ArcFace &
            \footnotesize CosFace &
            \footnotesize ElasticFace &
            \footnotesize PartialFC &
            \footnotesize ArcFace &
            \footnotesize CosFace &
            \footnotesize ElasticFace &
            \footnotesize PartialFC \\
            \midrule
            NNS &
            $8.61 \pm 0.02$ &
            $8.59 \pm 0.02$ &
            $8.60 \pm 0.02$ &
            $8.52 \pm 0.02$ &
            $3.12 \pm 0.01$ &
            $3.04 \pm 0.01$ &
            $3.13 \pm 0.01$ &
            $3.10 \pm 0.01$  \\
            $k$-means &
            $7.73 \pm 0.06$ &
            $7.73 \pm 0.06$ &
            $7.66 \pm 0.06$ &
            $7.61 \pm 0.06$ &
            $3.01 \pm 0.06$ &
            $2.81 \pm 0.07$ &
            $2.96 \pm 0.07$ &
            $3.01 \pm 0.06$ \\
            sp$k$-means &
            $8.06 \pm 0.05$ &
            $8.07 \pm 0.04$ &
            $8.04 \pm 0.05$ &
            $7.97 \pm 0.05$ &
            $2.65 \pm 0.08$ &
            $2.38 \pm 0.08$ &
            $2.63 \pm 0.08$ &
            --- \\
            DBSCAN &
            $6.94 \pm 0.17$ &
            $7.17 \pm 0.14$ &
            $7.47 \pm 0.10$ &
            $7.50 \pm 0.19$ &
            $2.13 \pm 0.30$&
            $2.29 \pm 0.30$&
            $2.46 \pm 0.43$&
            --- \\
            \midrule
            \textbf{LatentAlign} &
            \textbf{6.38} $\pm$ \textbf{0.02} &
            \textbf{6.37} $\pm$ \textbf{0.02} &
            \textbf{6.53} $\pm$ \textbf{0.02} &
            \textbf{6.25} $\pm$ \textbf{0.03} &
            \textbf{1.86} $\pm$ \textbf{0.02} & 
            \textbf{1.52} $\pm$ \textbf{0.02} &
            \textbf{1.84} $\pm$ \textbf{0.02} &
            \textbf{1.54} $\pm$ \textbf{0.03} \\
        \end{tabular}
    }
\end{table}

\subsection{Bias Discovery: Exposing Intersectional Vulnerabilities}
\label{Sec: Bias Evaluation}

In the previous section, we established that LatentAlign constructs highly coherent semantic subpopulations. We now demonstrate why this geometric precision matters for fairness and biometric security: highly coherent subpopulations expose the specific, intersectional failure modes of face recognition models that other methods miss.

\subsubsection{Evaluation Protocol for Worst-Case Auditing.} In unsupervised bias discovery, aggregating the error across all discovered clusters is misleading. Standard clustering algorithms produce thousands of groups, the vast majority of which represent common, low-error, mixed phenotypes. Auditing, however, is fundamentally a search for "worst-case" vulnerabilities. Therefore, we focus our parametric analysis on the top-100 most severe subpopulations (ranked by intra-group impostor similarity). This ensures we are comparing the algorithms exactly where it matters: their ability to find coherent groups with high error.

To ensure comparability across methods with very different clustering behaviors, we adopt the following evaluation protocol. For each algorithm, we run all combinations of parameters, collect all discovered groups, and bin groups by their size. Within each size bin, we rank groups by FMR and report the top-100 groups in each size. This controls for the fact that clustering algorithms can produce very large or very small clusters, which can trivially improve or worsen these metrics.

\begin{figure}[t]
    \centering
    \includegraphics[width=\linewidth]{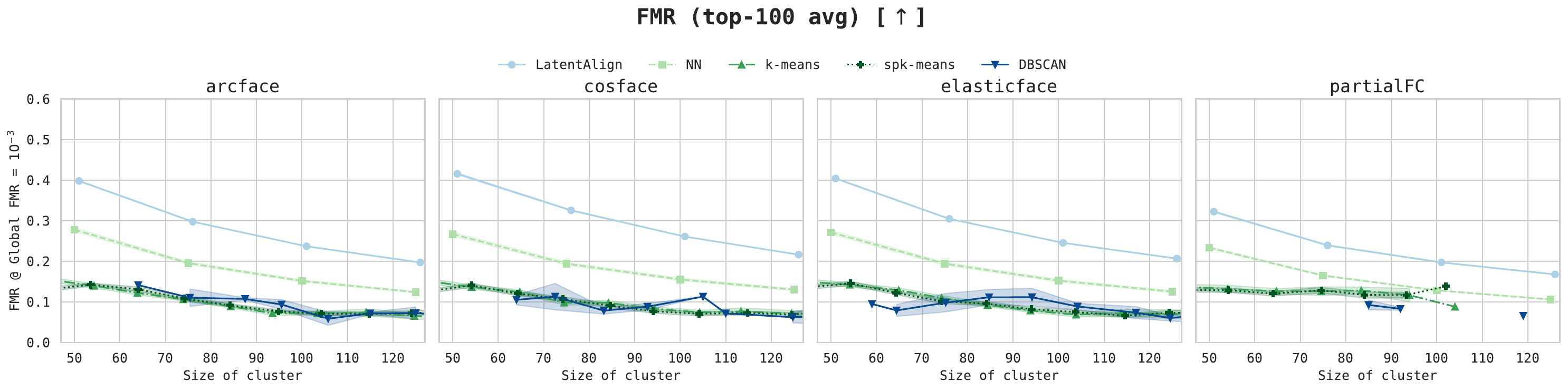}
    \caption{\textbf{FMR (False Match Rate) of groups discovered by each algorithm.} For each algorithm and FR backbone, we aggregate all discovered groups from all hyperparameter settings, bin them by group size, and compute the average FMR @ Global FMR=$10^{-3}$ of the top-100 most biased groups (higher = more bias revealed). LatentAlign consistently identifies groups with higher bias.}
    \label{fig:fmr_comparison}
\end{figure}

We compute error rates from impostor and genuine comparisons. In realistic security deployments, the FMR (False Match Rate) represents the primary operational risk, as false matches can lead to severe security or privacy breaches. Furthermore, in a discovered group of size $N=100$, the number of genuine pairs is extremely small (often close to zero if every image represents a different identity). This makes the FNMR (False Non-Match Rate) and EER (Equal Error Rate) statistically unstable and unreliable to estimate. Conversely, the same group contains up to 4,950 possible impostor pairs. This provides sufficient statistical power to compute a robust, group-specific FMR. Following standard real-world operating conditions, we evaluate the subpopulation FMR against a globally fixed threshold—specifically, the threshold at which the overall dataset achieves an FMR of $10^{-3}$.

As shown in Figure \ref{fig:fmr_comparison}, LatentAlign consistently discovers the most failure-prone subpopulations across all four face recognition models; i.e., the groups with the highest FMR, exposing intersectional groups that are more strongly biased than those found by standard baselines.

\subsubsection{Surpassing Supervised Annotations.} Traditional supervised auditing dilutes bias through demographic averaging. Current fairness benchmarks evaluate models across broad, predefined categories (e.g., "Asian" or "Female"). However, algorithmic bias is rarely uniform across a demographic; it is highly intersectional.

We evaluate whether the subpopulations discovered by LatentAlign correspond to groups with higher error rates, and whether these trends are consistent with groups defined using explicit attribute annotations. We use the RFW database given that it was specifically designed to measure bias (race bias).

Table~\ref{tab:FMRfixed} reports FMR (False Match Rate) at a fixed global threshold where FMR = $10^{-3}$, to compare groups discovered by LatentAlign against groups defined using explicit attribute annotations. Across all four backbones, LatentAlign routinely isolates subgroups exhibiting 1.4x to 4x higher FMR than those formed using predefined annotations. Precisely because the exact groups do not fully align, due to limitations in attribute coverage (e.g., no attribute for black-and-white photos), our method tends to uncover even more homogeneous groups, always resulting in higher FMR. Please refer to the Appendix \ref{Appendix: Evaluation} for the complete bias evaluation metrics and corresponding confidence intervals.

\begin{table}[t]
    \caption{\textbf{Comparison between LatentAlign and attribute annotations.} FMR (False Match Rate) at Global FMR = $10^{-3}$ for several groups discovered by LatentAlign and their corresponding groups defined using explicit attribute annotations. Each group  (A, B, C and D) includes the number of images and comparisons.}
    \label{tab:FMRfixed}
    \centering
    \resizebox{\columnwidth}{!}{%
        \begin{tabular}{l@{\hspace{8pt}} *{4}{c@{\hspace{4pt}}c@{\hspace{8pt}}}}
            \toprule
            \multirow{2}{*}{\textbf{Model}} & \multicolumn{2}{c}{\textbf{Group A} (130 / 8k)} & \multicolumn{2}{c}{\textbf{Group B} (143 / 10k)} & \multicolumn{2}{c}{\textbf{Group C} (96 / 4k)} & \multicolumn{2}{c}{\textbf{Group D} (100 / 5k)} \\
            \cmidrule(r){2-3}\cmidrule(r){4-5}\cmidrule(r){6-7}\cmidrule(r){8-9}
            & \textbf{Annotated} & \textbf{LatentAlign} & \textbf{Annotated} & \textbf{LatentAlign} & \textbf{Annotated} & \textbf{LatentAlign} & \textbf{Annotated} & \textbf{LatentAlign} \\
            \midrule
            ArcFace     & 0.015 & 0.052 ($3.4\times$) & 0.052 & 0.180 ($3.4\times$) & 0.152 & 0.215 ($1.4\times$) & 0.145 & 0.318 ($2.2\times$) \\
            CosFace     & 0.019 & 0.052 ($2.7\times$) & 0.068 & 0.245 ($3.6\times$) & 0.138 & 0.196 ($1.4\times$) & 0.117 & 0.279 ($2.3\times$) \\
            ElasticFace & 0.019 & 0.057 ($3.0\times$) & 0.052 & 0.167 ($3.2\times$) & 0.188 & 0.285 ($1.5\times$) & 0.175 & 0.436 ($2.4\times$) \\
            PartialFC   & 0.005 & 0.016 ($3.2\times$) & 0.024 & 0.106 ($4.4\times$) & 0.064 & 0.081 ($1.2\times$) & 0.053 & 0.140 ($2.6\times$) \\
            \bottomrule
        \end{tabular}
    }
\end{table}

\section{Discussion}

LatentAlign relies on the existence of a coherent underlying factor shared across a group of images. When such a factor is present, the latent direction aggregates and reflects the common representation, yielding a semantically interpretable vector. Conversely, if the images are unrelated, the latent direction captures inconsistent signals, resulting in noise and an incoherent mixture of faces without a distinct representation. In this case, projections become uninformative, and the resulting groups lose interpretability.

\subsection{Heuristic Seeding as a Lower Bound for Performance}

The quality of discovered subpopulations depends on the initial seed $\mathcal{S}$; a sensitivity analysis is provided in Appendix \ref{Appendix: Ablation Study} in the supplementary material. In our experiments, we relied on a simple similarity-graph heuristic to generate these seeds to maintain computational efficiency. Despite this rudimentary initialization, LatentAlign significantly outperformed standard clustering baselines across all metrics and model backbones.

Consequently, we view our reported performance as a lower bound. The theoretical ideal for LatentAlign is the exhaustive enumeration of small subsets, which would guarantee the discovery of all possible semantic directions without relying on local density heuristics. Future work could investigate optimized or "brute-force" initialization strategies, which we hypothesize would yield even sharper intersectional groups and expose deeper vulnerabilities within the embedding space.

\subsection{Limitations}

\textbf{Limited Attribute Coverage}
Although we validated robustness with annotations, the absence of certain attributes (e.g., black-and-white images) limits the completeness of comparisons. Extending the annotation coverage could provide a fuller picture of its potential.

\textbf{Observational Nature and Confounders.}
Our method is observational, relying solely on naturally occurring data. As such, it cannot disentangle semantic attributes from nuisance factors such as illumination, pose, or resolution. This limitation is inherent to any bias analysis performed without controlled interventions. 

\textbf{Linear Approximation of the Latent Space.} 
LatentAlign assumes locally linear directions in a highly dimensional embedding space. This linearization may overlook non-linear trajectories in the manifold, leading to partial or imperfect attribute representation.

\section{Conclusion}

In this work, we introduced LatentAlign, a simple yet effective method for discovering semantically coherent and interpretable subpopulations directly from face recognition embeddings. Unlike traditional clustering methods that rely purely on local distance metrics, LatentAlign exploits latent directions in the embedding space, leading to groups that are both more coherent and more interpretable. Our experimental evaluation across CelebA and RFW demonstrates three main contributions.

First, we demonstrated that the directions discovered by LatentAlign correspond to meaningful semantic attributes, as evidenced by both qualitative traversals and quantitative attribute probability trends using Arc2Face and InfiniteYou (see Section \ref{Sec: Interpretability}). These results highlight the interpretability of the latent space, even when derived from discriminative embeddings not trained for generative editing (such as GANs or autoencoders). Second, we showed that LatentAlign consistently produces groups with higher semantic coherence than $k$-means, spherical $k$-means, DBSCAN and nearest-neighbor search, across four state-of-the-art recognition models (ArcFace, CosFace, ElasticFace, PartialFC) and two widely used benchmarks (RFW, CelebA) (see Section \ref{Sec: Semantic Coherence} and Appendix~ \ref{Appendix: Quantitative Ablation}). Third, we confirm that the subpopulations identified by LatentAlign exhibit higher error rates consistent with trends observed using explicit attribute annotations (see Section \ref{Sec: Bias Evaluation} and Appendix~\ref{Appendix: Evaluation}).

Overall, LatentAlign offers a practical and interpretable method for analyzing embedding spaces, providing critical insights into both semantic structure and hidden biases. Our findings show that analyzing the latent geometry of these models provides a powerful, label-free approach for uncovering systematic vulnerabilities. We hope this work encourages further exploration of latent-structure-based fairness analysis in representation learning.

%
%
\bibliographystyle{splncs04}
\bibliography{refs}

\clearpage  

\appendix

\section{RFW Annotation and Validation} \label{Appendix: Labels}

\begin{table}[h]
    \centering
    \caption{OpenCompass VLM evaluation results of our selected Visual Language Models (VLMs) for annotating RFW. * Denotes the score of the unquantized version.}
    \label{tab:vlm-comparison}
    \resizebox{\columnwidth}{!}{%
        \begin{tabular}{l r l l r@{}}
            \toprule
             \textbf{Method} & \textbf{Params (B)} & \textbf{Language Model} & \textbf{Vision Model} & \textbf{Avg. Score} \\
            \midrule
             \href{https://huggingface.co/OpenGVLab/InternVL3-38B}{InternVL3-38B-AWQ}& 38.4  & Qwen2.5-32B     & InternViT-6B-v2.5 & 77.8*\\
             \href{https://huggingface.co/AIDC-AI/Ovis2-34B}{Ovis2-34B-GPTQ-Int8}& 34.9  & Qwen2.5-32B     & AIMv2-1B          & 76.5*\\
             \href{https://huggingface.co/BytedanceDouyinContent/SAIL-VL-1d6-8B}{SAIL-VL-1.6-8B} & 8.33  & Qwen2.5-7B      & AIMv2 Huge        & 73.6\\
             \href{https://huggingface.co/THUdyh/Ola-7b}{Ola-7b}                       & 8.88  & Qwen2.5-7B      & Oryx-ViT          & 72.6 \\
             \href{https://huggingface.co/Qwen/Qwen2.5-VL-32B-Instruct-AWQ}{Qwen2.5-VL-32B-AWQ}& 32.5 & Qwen2.5-32B & QwenViT & -\\
            \bottomrule
        \end{tabular}
    }
\end{table}

Given the noise of CelebA attribute annotations, we decided to provide more robust evidence of the better clustering. For that we annotated RFW with five state of the art VLMs: InternVL3-38B-AWQ \cite{chen2024internvl}, Ola-7b \cite{liu2025ola}, Ovis2-34B-GPTQ-Int8 \cite{lu2024ovis}, Qwen2.5-VL-32B-Instruct-AWQ \cite{Qwen2.5-VL}, and SAIL-VL-1d6-8B \cite{dong2025scalable}.

The models were chosen using the benchmark on the OpenCompass VLM leaderboard\footnote{\url{https://huggingface.co/spaces/opencompass/open_vlm_leaderboard}}. The top 20 open source models belong to the 5 VLM families, and they vary in size. We chose one model from each family, 3 being of size $\sim$32B, and 2 of size 7B. The 32B models are quantized versions of the models (to fit a GPU). Table \ref{tab:vlm-comparison} shows the models' vision and language backbones, number of parameters, and average score. The Qwen2.5-VL-32B model does not appear in the leaderboard.

The following prompt was used to elicit facial attribute classification from the VLMs:

\begin{lstlisting}[style=promptstyle]
You are an image analysis tool specialized in facial attribute classification. 
For the provided face image, output a JSON object with the following attributes:

{
  @k"gender"@: [@v"male"@, @v"female"@],
  @k"age"@: [@v"young"@, @v"middle-aged"@, @v"senior"@],
  @k"skin_color"@: [@v"light"@, @v"medium"@, @v"dark"@],
  @k"ancestry"@: [@v"asian"@, @v"south_asian"@, @v"black"@, @v"latino/hispanic"@, @v"middle_eastern"@, 
                    @v"white"@, @v"indigenous"@],
  @k"hair_color"@: [@v"black"@, @v"brown"@, @v"red"@, @v"blonde"@, @v"gray"@, @v"other"@],
  @k"bangs"@: [@v"yes"@, @v"no"@],
  @k"bald"@: [@v"yes"@, @v"no"@],
  @k"beard"@: [@v"no"@, @v"mustache"@, @v"stubble"@, @v"full"@],
  @k"glasses"@: [@v"no"@, @v"regular"@, @v"sun"@],
  @k"headwear"@: [@v"no"@, @v"beanie"@, @v"cap"@, @v"hat"@, @v"headband"@, @v"hijab"@, @v"helmet"@, @v"turban"@]
}

Ensure the labeling is based on visible evidence only. If an attribute is unclear, 
return "unknown".

Only output the JSON without any additional explanation or text.
Example JSON output:
{
  @k"gender"@: @v"female"@,
  @k"age"@: @v"middle-aged"@,
  @k"skin_color"@: @v"light"@,
  @k"ancestry"@: @v"asian"@,
  @k"hair_color"@: @v"black"@,
  @k"bangs"@: @v"no"@,
  @k"bald"@: @v"no"@,
  @k"beard"@: @v"no"@,
  @k"glasses"@: @v"sun"@,
  @k"headwear"@: @v"beanie"@
}
\end{lstlisting}

Table \ref{tab:rfw_annotations} breaks down each attribute annotation showing how many samples fall into each class, both as a raw count and as a percentage of the total dataset. There is also an "Unknown" category for each attribute, which indicates how many images had no consensus among the models for that particular attribute. We quantify labeling bias computing compute agreement scores. These scores reflect the proportion of VLMs that agreed on the final label. For example, for the final label "mustache" in the 'beard' attribute of an image, if 3 VLMs say 'mustache', 1 says 'stubble' and 1 says 'no', the agreement ratio for that image and that attribute is 3/5 = 0.6. Table \ref{tab:rfw_annotations} presents mean agreement and standard deviation by attribute. Most classes show a high degree of agreement. Only \textit{Hair Color} and \textit{Ancestry} show less agreement for less frequent classes like \textit{Latino/Hispanic}, \textit{Indigenous}, or \textit{Other}.

\begin{table}[!ht]
  \centering
  \caption{Attribute annotations and agreement scores (mean and standard deviation) of the RFW dataset using five state-of-the-art Visual Language Models. \textit{Unknown} class means there was no consensus among models. Total samples: 40,607. }
  \label{tab:rfw_annotations}
  \scriptsize
  \begin{tabular}{llrrcc}
    \textbf{Category} & \textbf{Class} & \textbf{Count} & \textbf{\%} & \textbf{Mean Agr.} & \textbf{Std Agr.}\\
    \midrule

    \multirow{4}{*}{Age}
      & Middle-aged & 19,072 & 47.0 & 0.88 & 0.15 \\
      & Young       & 16,144 & 39.8 & 0.90 & 0.15 \\
      & Senior      & 5,312  & 13.1 & 0.92 & 0.14 \\
      & \tiny\textcolor{gray}{\textit{Unknown}} & \tiny\textcolor{gray}{79} & \tiny\textcolor{gray}{0.2} & & \\
    \midrule

    \multirow{8}{*}{Ancestry}
      & Black          & 10,396 & 25.6 & 0.99 & 0.04 \\
      & White          & 10,161 & 25.0 & 0.95 & 0.11 \\
      & Asian          & 9,547  & 23.5 & 0.98 & 0.07 \\
      & South Asian    & 9,191  & 22.6 & 0.93 & 0.12 \\
      & Latino/Hispanic& 379    & 0.9  & 0.67 & 0.10 \\
      & Middle Eastern & 146    & 0.4  & 0.70 & 0.12 \\
      & Indigenous     & 18     & 0.0  & 0.81 & 0.18 \\
      & \tiny\textcolor{gray}{\textit{Unknown}} & \tiny\textcolor{gray}{769} & \tiny\textcolor{gray}{1.9} & & \\
    \midrule

    \multirow{3}{*}{Bald}
      & No  & 37,553 & 92.5 & 0.98 & 0.07 \\
      & Yes & 3,047  & 7.5  & 0.84 & 0.15 \\
      & \tiny\textcolor{gray}{\textit{Unknown}} & \tiny\textcolor{gray}{7} & \tiny\textcolor{gray}{0.0} & & \\
    \midrule

    \multirow{3}{*}{Bangs}
      & No  & 35,280 & 86.9 & 0.94 & 0.11 \\
      & Yes & 5,297  & 13.0 & 0.86 & 0.16 \\
      & \tiny\textcolor{gray}{\textit{Unknown}} & \tiny\textcolor{gray}{30} & \tiny\textcolor{gray}{0.1} & & \\
    \midrule

    \multirow{5}{*}{Beard}
      & No       & 30,356 & 74.8 & 0.91 & 0.12 \\
      & Stubble  & 3,940  & 9.7  & 0.74 & 0.15 \\
      & Mustache & 3,677  & 9.1  & 0.89 & 0.16 \\
      & Full     & 1,263  & 3.1  & 0.77 & 0.16 \\
      & \tiny\textcolor{gray}{\textit{Unknown}} & \tiny\textcolor{gray}{1371} & \tiny\textcolor{gray}{3.4} & & \\
    \midrule

    \multirow{3}{*}{Gender}
      & Male   & 30,191 & 74.3 & 1.00 & 0.04 \\
      & Female & 10,400 & 25.6 & 0.99 & 0.06 \\
      & \tiny\textcolor{gray}{\textit{Unknown}} & \tiny\textcolor{gray}{16} & \tiny\textcolor{gray}{0.0} & & \\
    \midrule

    \multirow{4}{*}{Glasses}
      & No      & 34,589 & 85.2 & 1.00 & 0.03 \\
      & Regular & 4,914  & 12.1 & 0.98 & 0.08 \\
      & Sun     & 1,080  & 2.7  & 0.94 & 0.12 \\
      & \tiny\textcolor{gray}{\textit{Unknown}} & \tiny\textcolor{gray}{24} & \tiny\textcolor{gray}{0.1} & & \\
    \midrule

    \multirow{7}{*}{Hair Color}
      & Black  & 26,114 & 64.3 & 0.96 & 0.11 \\
      & Brown  & 5,795  & 14.3 & 0.82 & 0.15 \\
      & Gray   & 5,529  & 13.6 & 0.91 & 0.14 \\
      & Blonde & 1,780  & 4.4  & 0.92 & 0.14 \\
      & Red    & 935    & 0.8  & 0.84 & 0.16 \\
      & Other  & 129    & 0.3  & 0.65 & 0.06 \\
      & \tiny\textcolor{gray}{\textit{Unknown}} & \tiny\textcolor{gray}{942} & \tiny\textcolor{gray}{2.3} & & \\
    \midrule

    \multirow{9}{*}{Headwear}
      & No        & 35,543 & 87.5 & 0.99 & 0.04 \\
      & Cap       & 1,991  & 4.9  & 0.92 & 0.14 \\
      & Hat       & 1,089  & 2.7  & 0.87 & 0.17 \\
      & Headband  & 513    & 1.3  & 0.81 & 0.16 \\
      & Beanie    & 304    & 0.7  & 0.89 & 0.16 \\
      & Turban    & 262    & 0.6  & 0.94 & 0.13 \\
      & Helmet    & 255    & 0.6  & 0.89 & 0.14 \\
      & Hijab     & 199    & 0.5  & 0.91 & 0.15 \\
      & \tiny\textcolor{gray}{\textit{Unknown}} & \tiny\textcolor{gray}{451} & \tiny\textcolor{gray}{1.1} & & \\
    \midrule

    \multirow{4}{*}{Skin Tone}
      & Medium & 15,499 & 38.2 & 0.76 & 0.12 \\
      & Light  & 13,701 & 33.7 & 0.91 & 0.15 \\
      & Dark   & 10,952 & 27.0 & 0.97 & 0.09 \\
      & \tiny\textcolor{gray}{\textit{Unknown}} & \tiny\textcolor{gray}{455} & \tiny\textcolor{gray}{1.1} & & \\
  \end{tabular}
\end{table}

\subsection*{Annotation Merging via Majority Voting}

To merge the predictions of the five VLMs into a single robust annotation per attribute, we used a majority voting scheme that explicitly ignores \texttt{unknown} values. For each image and attribute, we collected the predictions from all five models and discarded any labeled as \texttt{unknown}. The final consensus label was defined as the label that received strictly more than half of the valid (i.e., non-\texttt{unknown}) predictions. If no such label existed—either due to a tie or insufficient agreement—the annotation was marked as \texttt{unknown}.

Formally, let $A_i^j$ be the label predicted by model $i$ for attribute $j$, and define $V = \{A_i^j \mid A_i^j \neq \texttt{unknown}\}$ as the set of valid predictions for that attribute. Let $n = |V|$ denote the number of valid votes. The consensus label $C^j$ is then given by:

\[
C^j = 
\begin{cases}
\ell \in V, & \text{if } \mathrm{count}(\ell) > \frac{n}{2} \\
\texttt{unknown}, & \text{otherwise}
\end{cases}
\]

This procedure ensures that only strong and unambiguous agreement among the models leads to an assigned label, while also allowing for occasional uncertainty (expressed via \texttt{unknown}) improving the reliability of the annotations used in our clustering analysis. We emphasize that the labels obtained via majority voting among VLMs serve as weak annotations and are not ground truth. In rare cases (fewer than 0.1\% of samples), images with multiple faces were manually disambiguated by selecting the most centrally located face.

\subsection*{Human Annotation Validation}

We validated the automatic attribute annotations against a human-annotated sample. Table \ref{tab:human_validation} shows per-attribute raw match rate (the fraction of exact label matches) and Cohen’s $\kappa$ (agreement corrected for chance). The validation sample contains 1,000 matched images per attribute. For binary attributes with strong class balance (e.g., gender, glasses), raw accuracy and $\kappa$ are both very high (\textit{gender}: 99\% match, $\kappa$ = 0.974; \textit{glasses}: 99\% match, $\kappa$ = 0.969), indicating near-perfect agreement. Multi-class attributes (\textit{age}, \textit{skin color}, \textit{ancestry}, \textit{hair color}) also show very strong agreement (e.g., \textit{ancestry} $\kappa$ = 0.931), indicating the automatic labels are reliable for downstream analyses.

Cohen’s $\kappa$ measures inter-rater reliability for qualitative (categorical) classes and controls for chance agreement and class imbalance; it should be preferred to raw match rate when classes are uneven. Per-class metrics (precision/recall/F1) and confusion matrices (provided in the supplement) reveal that some low-support classes (for example, the \textit{Latino/Hispanic} ancestry class and several rare \textit{Hair-color} categories) have poor per-class precision/recall despite high overall accuracy. These small-class failures do not substantially affect the aggregate statistics but are important to note when analyzing model behaviour for specific subgroups.

\begin{figure}[t]
    \centering
    \begin{subfigure}{0.2\linewidth}
        \centering
        \includegraphics[width=\linewidth]{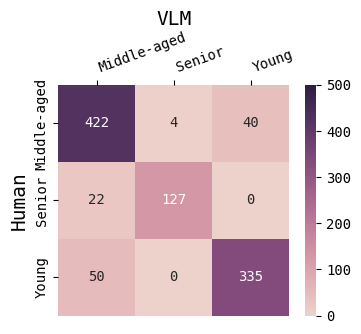}
        \caption{Age}
    \end{subfigure}
    \begin{subfigure}{0.22\linewidth}
        \centering
        \includegraphics[width=\linewidth]{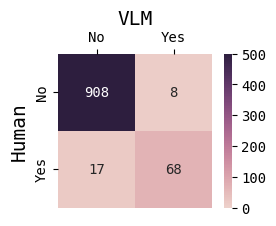}
        \caption{Bald}
    \end{subfigure}
    \begin{subfigure}{0.22\linewidth}
        \centering
        \includegraphics[width=\linewidth]{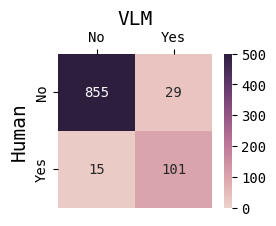}
        \caption{Bangs}
    \end{subfigure}
    \begin{subfigure}{0.22\linewidth}
        \centering
        \includegraphics[width=\linewidth]{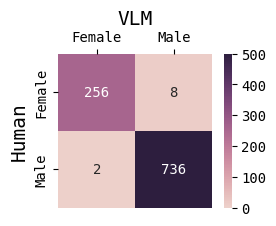}
        \caption{Gender}
    \end{subfigure}
    \centering
    \begin{subfigure}{0.29\linewidth}
        \centering
        \includegraphics[width=\linewidth]{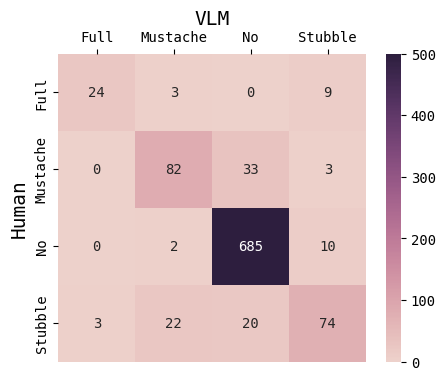}
        \caption{Beard}
    \end{subfigure}
    \begin{subfigure}{0.3\linewidth}
        \centering
        \includegraphics[width=\linewidth]{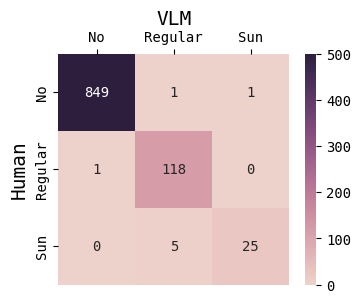}
        \caption{Glasses}
    \end{subfigure}
    \begin{subfigure}{0.3\linewidth}
        \centering
        \includegraphics[width=\linewidth]{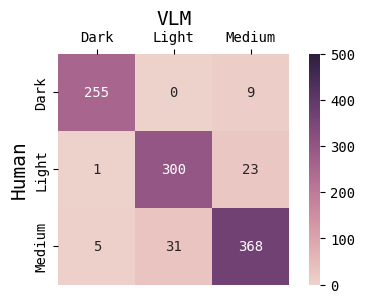}
        \caption{Skin Color}
    \end{subfigure}
    \centering
    \begin{subfigure}{0.45\linewidth}
        \centering
        \includegraphics[width=\linewidth]{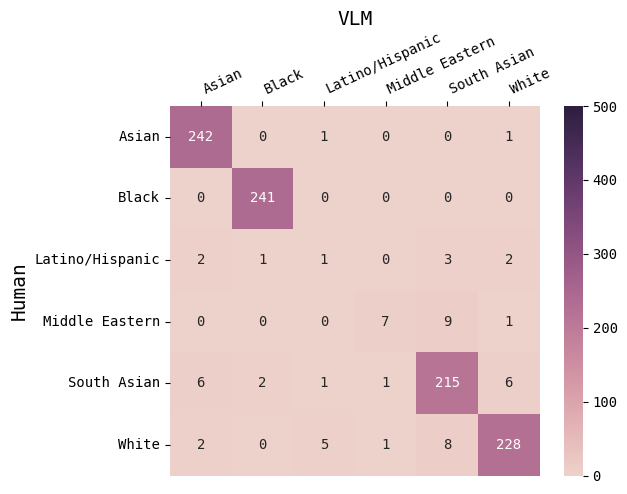}
        \caption{Ancestry}
    \end{subfigure}
    \begin{subfigure}{0.37\linewidth}
        \centering
        \includegraphics[width=\linewidth]{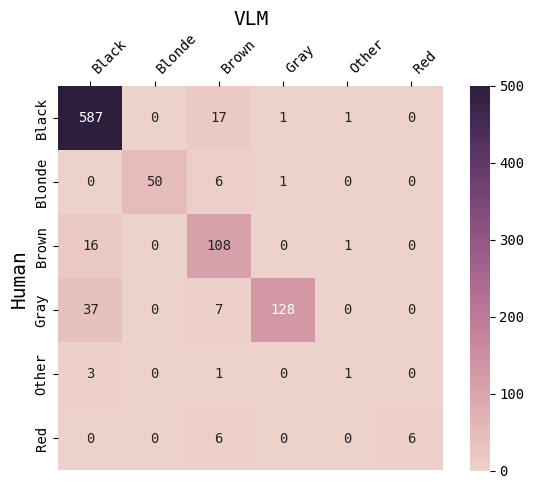}
        \caption{Hair Color}
    \end{subfigure}
    
    \centering
    \begin{subfigure}{0.45\linewidth}
        \centering
        \includegraphics[width=\linewidth]{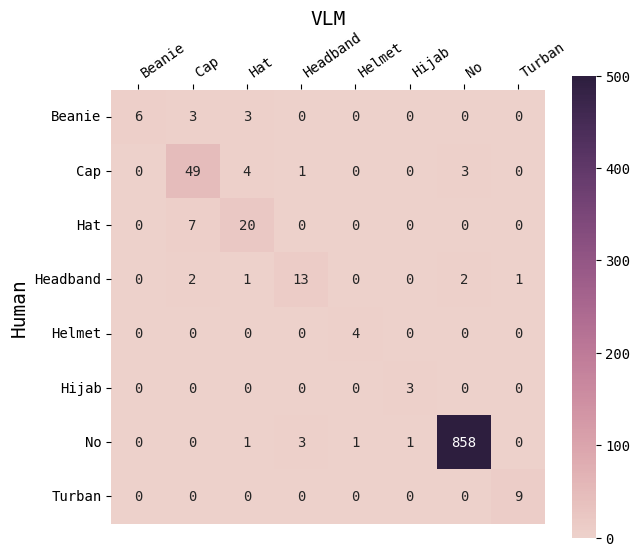}
        \caption{Headwear}
    \end{subfigure}
    \caption{Confusion matrices for each attribute between human annotations and VLM annotations.}
    \label{fig:confusion_matrices}
\end{figure}

The low agreement for the attribute \textit{Beard} is due to the fact that the human annotators considered goatees and other beard styles fall into the \textit{Stubble} category, while VLM did not. Figure \ref{fig:confusion_matrices} include the full confusion matrices of subcategory-level matching.

\begin{table}[t]
    \centering
    \caption{\textbf{Validation of automatic attribute annotations against a human-annotated 1K sample.} Each row reports the Match Rate (fraction of exact matches) and Cohen's $\kappa$ (agreement corrected for chance). High $\kappa$ values indicate substantial-to-almost-perfect agreement for most attributes.}
    \label{tab:human_validation}
    \begin{tabular}{lcc}
        Attribute & Match Rate ($\uparrow$) & Cohen's $\kappa$ ($\uparrow$) \\
        \midrule
        Age         & 0.8840 & 0.8085 \\
        Ancestry    & 0.9473 & 0.9306 \\
        Bald        & 0.9750 & 0.8312 \\
        Bangs       & 0.9560 & 0.7961 \\
        Beard       & 0.8918 & 0.7462 \\
        Gender      & 0.9900 & 0.9741 \\
        Glasses     & 0.9920 & 0.9694 \\
        Hair Color & 0.9007 & 0.8183 \\
        Headwear    & 0.9668 & 0.8630 \\
        Skin Color & 0.9304 & 0.8941 \\
    \end{tabular}
\end{table}

\clearpage

\section{Traversals} \label{Appendix: Traversals}

To reconstruct images from facial embeddings, we use Arc2Face \cite{papantoniou2024arc2face}, which synthesizes high-quality, identity-preserving images given ResNet-100 ArcFace \cite{deng2019arcface} embeddings, and InfiniteYou \cite{jiang2025infiniteyou}, from Improved ResNet-50 with Squeeze-and-Excitation.

Since face embeddings lie on the unit hypersphere, we cannot apply edits via direct vector addition. Instead, we use spherical linear interpolation (slerp) \cite{shoemake1985slerp} to interpolate between embeddings. This approach respects the geometry of the hypersphere and provides fine-grained control over the edit intensity.

In contrast to linear interpolation or simple vector averaging—which may produce off-manifold results or require manual tuning of edit strength depending on the embedding and direction—slerp offers a principled mechanism for traversal along geodesics.

The slerp between two unit vectors $p_0$ and $p_1$ is defined as:

\[
\text{Slerp}(p_0, p_1; t) = \frac{\sin((1 - t) \theta)}{\sin \theta} p_0 + \frac{\sin(t \theta)}{\sin \theta} p_1
\]

where $ t\in [0,1]$  is the interpolation parameter and $\theta = \cos^{-1}(p_0 \cdot p_1)$ is the angle between $p_0$ and $p_1$. The interpolation smoothly transitions from $p_0$ to $p_1$ along the great arc on the hypersphere.

For the results shown in Figure \ref{fig:subtraction}, we used a fixed interpolation strength of $t=-0.5$ to subtract and visualize the edit direction. For Figure \ref{fig:addition}, the interpolation strength had to be adjusted between $t=0.45$ and $t=0.5$ depending on the embedding and direction. In a few rare cases, a value as low as $t=0.4$ was required to ensure plausible and visually coherent outputs.

\section{Bias evaluation} \label{Appendix: Evaluation}

\begin{figure}[h]
    \centering
    \begin{subfigure}{0.49\linewidth}
        \centering
        \includegraphics[width=\linewidth]{distributions.png}
        \caption{Arcface}
        \label{fig:dist1}
    \end{subfigure}\hfill
    \begin{subfigure}{0.49\linewidth}
        \centering
        \includegraphics[width=\linewidth]{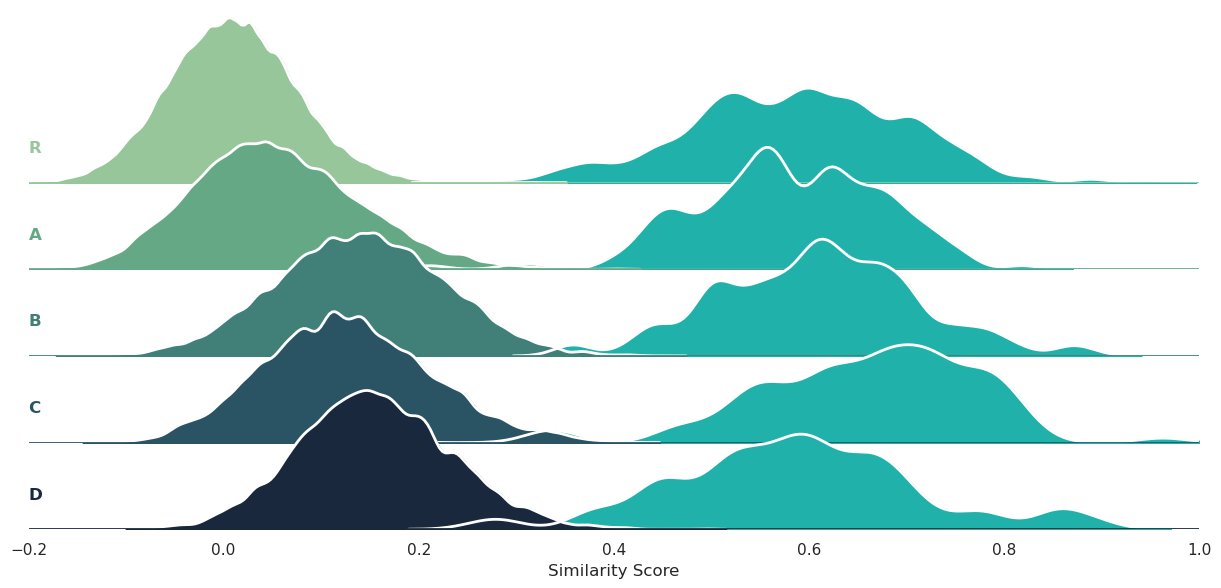}
        \caption{Cosface}
        \label{fig:dist2}
    \end{subfigure}
    \begin{subfigure}{0.49\linewidth}
        \centering
        \includegraphics[width=\linewidth]{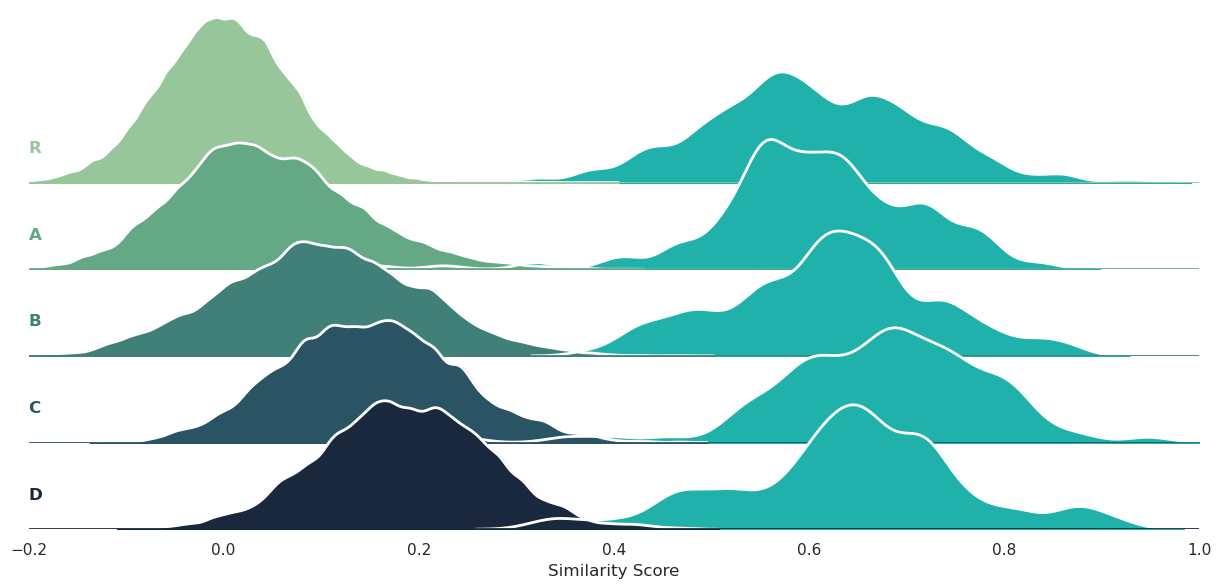}
        \caption{ElasticFace}
        \label{fig:dist3}
    \end{subfigure}\hfill
    \begin{subfigure}{0.49\linewidth}
        \centering
        \includegraphics[width=\linewidth]{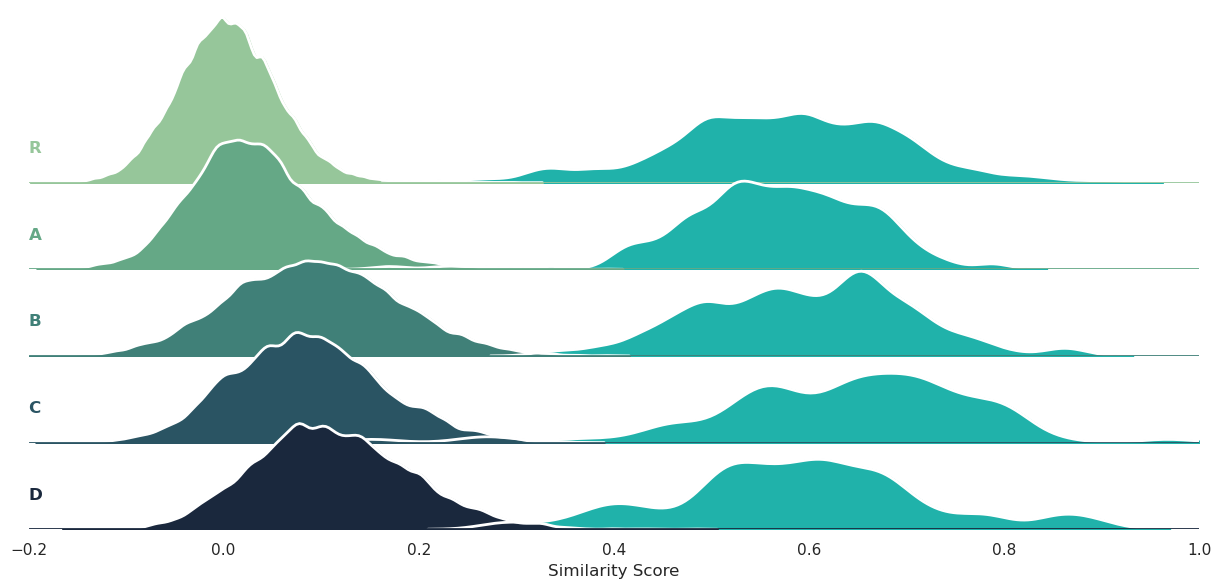}
        \caption{Partial FC}
        \label{fig:dist4}
    \end{subfigure}
    \caption{Distribution of cosine similarity scores for impostor and genuine comparisons within the same group.}
    \label{fig:all_dists}
\end{figure}

\begin{table}[ht]
  \centering
  \small
  \caption{Performance (EER and FNMR) of the face recognition models across the discovered groups (A-D), with standard deviation across groups (lower means fairer, less biased). Each group includes the number of images and comparisons.}
  \label{tab:bias-eval}
  \begin{tabular}{l *{4}{S[scientific-notation=true, table-format=1.2e-2]} S[table-format=1.4]}
    & \multicolumn{4}{c}{\textbf{EER}} & \\
    \cmidrule(lr){2-5}
    \textbf{Model} & {\textbf{Group A}} & {\textbf{Group B}} & {\textbf{Group C}} & {\textbf{Group D}} & {$\sigma$} \\
    & {(130 / 8k)} & {(143 / 10k)} & {(96 / 4k)} & {(100 / 5k)} & \\
    \midrule
    ArcFace     & 8.70e-03 & 4.00e-04 & 2.42e-02 & 1.00e-03 & 0.0096 \\
    CosFace     & 5.10e-03 & 6.40e-03 & 1.83e-02 & 1.57e-02 & 0.0057 \\
    ElasticFace & 8.20e-03 & 1.40e-03 & 1.22e-02 & 1.89e-02 & 0.0063 \\
    Partial FC  & 6.30e-03 & 1.20e-03 & 1.94e-02 & 1.78e-02 & 0.0077 \\
  \end{tabular}

  \vspace{0.75em}

  \begin{tabular}{l *{4}{S[scientific-notation=true, table-format=1.2e-2]} S[table-format=1.4]}
    & \multicolumn{4}{c}{\textbf{FNMR @ FMR=$10^{-2}$}} & \\
    \cmidrule(lr){2-5}
    \textbf{Model} & {\textbf{Group A}} & {\textbf{Group B}} & {\textbf{Group C}} & {\textbf{Group D}} & {$\sigma$} \\
    & {(130 / 8k)} & {(143 / 10k)} & {(96 / 4k)} & {(100 / 5k)} & \\
    \midrule
    ArcFace     & 1.15e-02 & 0.00e-00 & 2.08e-02 & 0.00e-00 & 0.0087 \\
    CosFace     & 5.70e-03 & 0.00e-00 & 3.12e-02 & 2.52e-02 & 0.0130 \\
    ElasticFace & 1.15e-02 & 0.00e-00 & 2.08e-02 & 3.36e-02 & 0.0123 \\
    Partial FC  & 5.70e-03 & 0.00e-00 & 2.08e-02 & 1.68e-02 & 0.0083 \\
  \end{tabular}

  \vspace{0.75em}

  \begin{tabular}{l *{4}{S[scientific-notation=true, table-format=1.2e-2]} S[table-format=1.4]}
    & \multicolumn{4}{c}{\textbf{FNMR @ FMR=$10^{-3}$}} & \\
    \cmidrule(lr){2-5}
    \textbf{Model} & {\textbf{Group A}} & {\textbf{Group B}} & {\textbf{Group C}} & {\textbf{Group D}} & {$\sigma$} \\
    & {(130 / 8k)} & {(143 / 10k)} & {(96 / 4k)} & {(100 / 5k)} & \\
    \midrule
    ArcFace     & 2.30e-02 & 0.00e+00 & 3.12e-02 & 4.20e-02 & 0.0154 \\
    CosFace     & 1.72e-02 & 1.97e-02 & 4.17e-02 & 7.56e-02 & 0.0234 \\
    ElasticFace & 2.30e-02 & 6.60e-03 & 3.12e-02 & 8.40e-02 & 0.0290 \\
    Partial FC  & 1.72e-02 & 6.60e-03 & 3.12e-02 & 8.40e-02 & 0.0297 \\
  \end{tabular}
\end{table}

\begin{table}[t]
  \caption{False Match Rates (FMR) with 95\% confidence intervals at Global FMR = $10^{-3}$ for each demographic group across four face recognition methods. Confidence intervals are computed via bootstrap resampling (1,000 iterations) at the image level.}
  \label{tab:FMRconfidence}
  \centering
  \small
  \resizebox{\textwidth}{!}{%
  \begin{tabular}{llllll}
    \textbf{Method} & \textbf{Group R} & \textbf{Group A} & \textbf{Group B} & \textbf{Group C} & \textbf{Group D} \\
    \midrule
    ArcFace     & 0.0057 ± 0.002 & 0.0522 ± 0.013 & 0.1798 ± 0.021 & 0.2167 ± 0.036 & 0.3181 ± 0.043 \\
    CosFace     & 0.0045 ± 0.002 & 0.0521 ± 0.014 & 0.2447 ± 0.027 & 0.1952 ± 0.038 & 0.2780 ± 0.036 \\
    ElasticFace & 0.0043 ± 0.002 & 0.0569 ± 0.015 & 0.1676 ± 0.021 & 0.2861 ± 0.044 & 0.4365 ± 0.050 \\
    PartialFC   & 0.0016 ± 0.001 & 0.0161 ± 0.007 & 0.1060 ± 0.016 & 0.0809 ± 0.022 & 0.1400 ± 0.033 \\
  \end{tabular}
  }
\end{table}

\begin{figure}[ht]
    \centering
    \begin{subfigure}{0.49\linewidth}
        \centering
        \includegraphics[width=\linewidth]{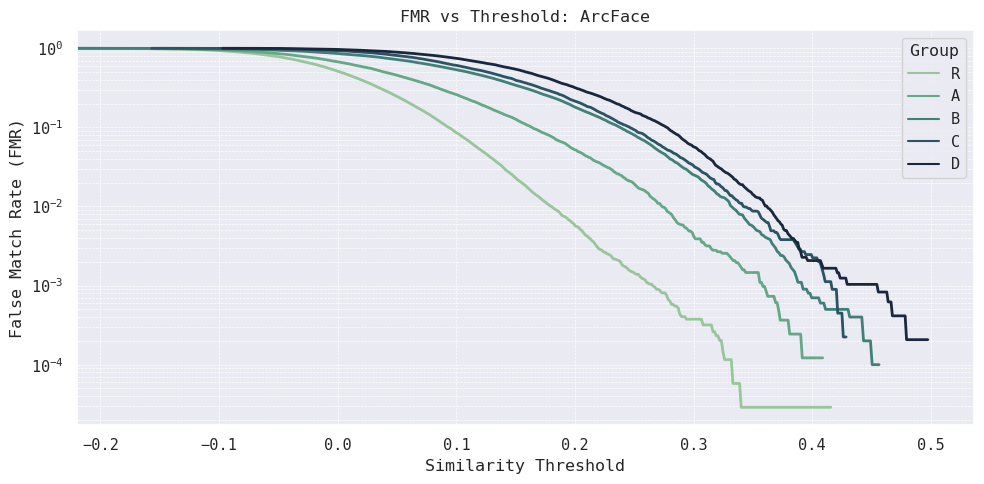}
        \caption{Arcface}
        \label{fig:fmr1}
    \end{subfigure}\hfill
    \begin{subfigure}{0.49\linewidth}
        \centering
        \includegraphics[width=\linewidth]{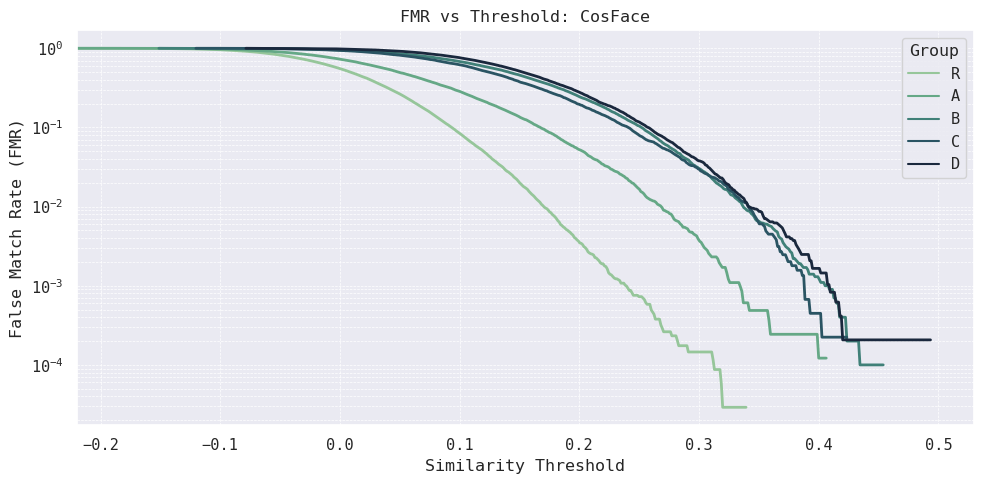}
        \caption{Cosface}
        \label{fig:fmr2}
    \end{subfigure}
    \begin{subfigure}{0.49\linewidth}
        \centering
        \includegraphics[width=\linewidth]{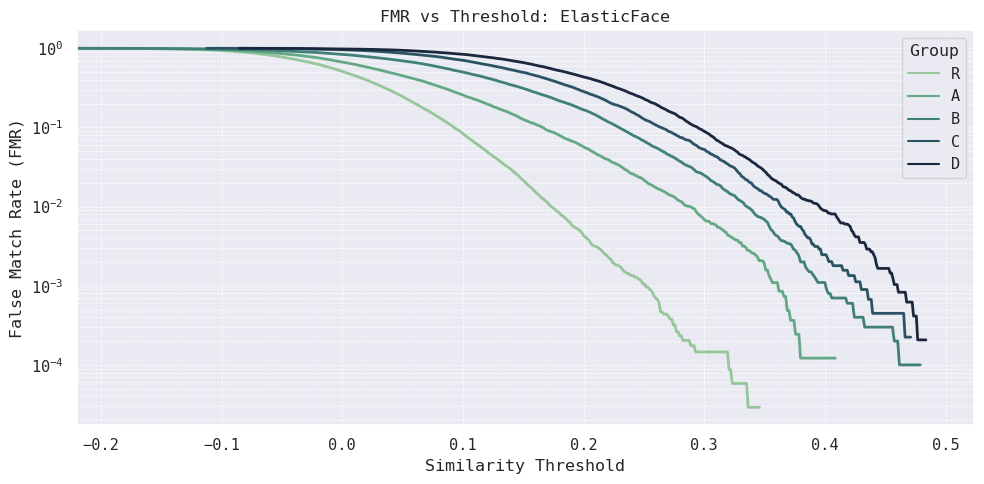}
        \caption{ElasticFace}
        \label{fig:fmr3}
    \end{subfigure}\hfill
    \begin{subfigure}{0.49\linewidth}
        \centering
        \includegraphics[width=\linewidth]{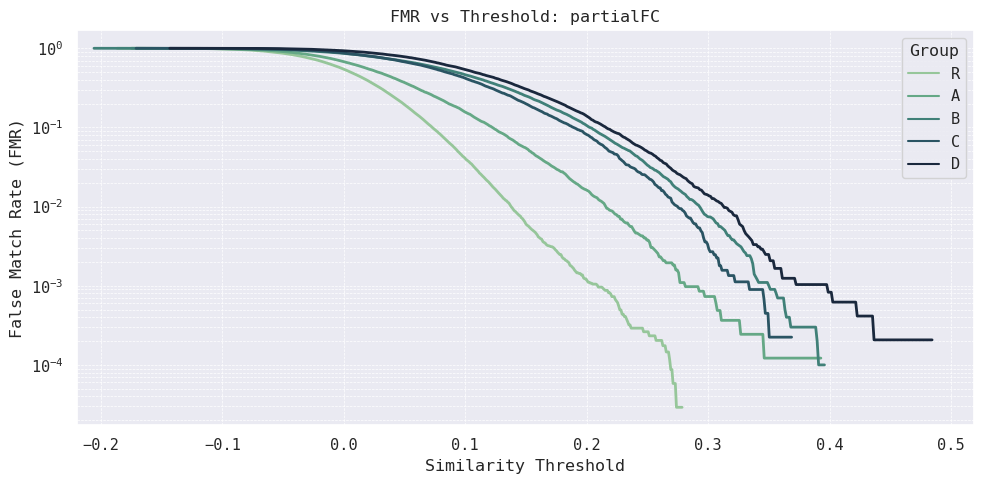}
        \caption{Partial FC}
        \label{fig:fmr4}
    \end{subfigure}
    \caption{False Match Rate (FMR) curves for each face recognition model for all groups.}
    \label{fig:all_fmr}
\end{figure}

We evaluate face recognition performance across some discovered groups (A-D) by forming genuine and imposter pairs exclusively within each group. Genuine pairs are pairs of images belonging to the same person and imposter pairs are formed by pairing images from different identities. Similarity scores are computed using the cosine similarity of the corresponding face embeddings.

Figure \ref{fig:all_dists} shows the distribution of cosine similarity scores per model for genuine and imposter pairs within each group. The right distributions (gray) corresponds to genuine pairs (higher similarity values), while the left distribution (colored) corresponds to impostor pairs (lower similarity values). Darker colors indicate groups with higher error (imposter distributions shifted towards genuine), while lighter colors indicate groups with lower error.

From the resulting score distributions, we compute:

\textbf{Equal Error Rate (EER)}: the operating point where false match rate (FMR) equals false non-match rate (FNMR).

\textbf{FNMR at FMR=$10^{-2}$ and $10^{-3}$}: the FNMR (False Non-Match rate) value when the decision threshold is set to yield an FMR (False Match Rate) of $10^{-2}$ and $10^{-3}$.

Table \ref{tab:bias-eval} summarizes the verification metrics for all groups. Sample sizes vary per group and are reported alongside EER, FNMR @ $10^{-2}$ FMR, and FNMR @ $10^{-3}$ FMR. The standard deviation ($\sigma$) between A-D groups is a common intuitive proxy for bias in biometric systems as it indicates the statistical measure of variance across groups  — higher $\sigma$ means more disparity.  FNMR @ $10^{-3}$ FMR means the false non-match rate at a fixed threshold for which the false match rate is $10^{-3}$. 

Although we observe that Groups C and D exhibit higher EER and FNMR than A and B, confirming the presence of bias in model performance for these groups, these groups lack the necessary number of comparisons for detailed measurements (notice the zeroes results for group B in FNMR @ FMR=$10^{-2}$). 

Although Groups C and D exhibit higher EER and FNMR compared to Groups A and B, indicating bias in the model's performance, they groups lack sufficient comparative data for detailed analysis at high error rates. For example, Group B has FNMR values of zero at FMR = $10^{-2}$. To address this limitation, Table~\ref{tab:FMRconfidence} reports false match rates (FMR) at a fixed global threshold where FMR = $10^{-3}$ with confidence intervals to assess statistical significance. Confidence intervals are computed via bootstrap resampling at the image level (1,000 iterations). 

When evaluating face recognition systems at a fixed FMR (False Match Rate) of $10^{-3}$, the decision threshold is chosen to allow only 1 in 1,000 impostor comparisons to be incorrectly accepted. The corresponding False Non-Match Rate (FNMR) is then computed over genuine comparisons using this threshold. However, when the number of genuine pairs is small—for example, approximately 120, like in our case—each false non-match has a disproportionately large effect on the FNMR estimate. Specifically, each false rejection increases the FNMR by roughly 0.83\% (i.e., 1/120).  This implies that the FNMR can only take on discrete values, such as 0\%, 0.83\%, 1.67\%, etc., making it difficult to make fine-grained measurements. As a result, observed differences in FNMR at low FMRs may not be statistically significant or practically meaningful. Small improvements reported at this operating point could simply be due to random variation rather than true differences in system performance.

A more robust analysis of imposter distributions (in the order of thousands of scores) is to compute the FMR curve group-wise as a function of the threshold. Figure \ref{fig:all_fmr} shows the FMR curves for each face recognition model—ArcFace, CosFace, ElasticFace, and Partial FC—disaggregated by group. These curves illustrate how the False Match Rate varies across groups with the decision threshold, highlighting disparities between different groups.

\begin{figure}[t]
    \centering
    \includegraphics[width=\linewidth]{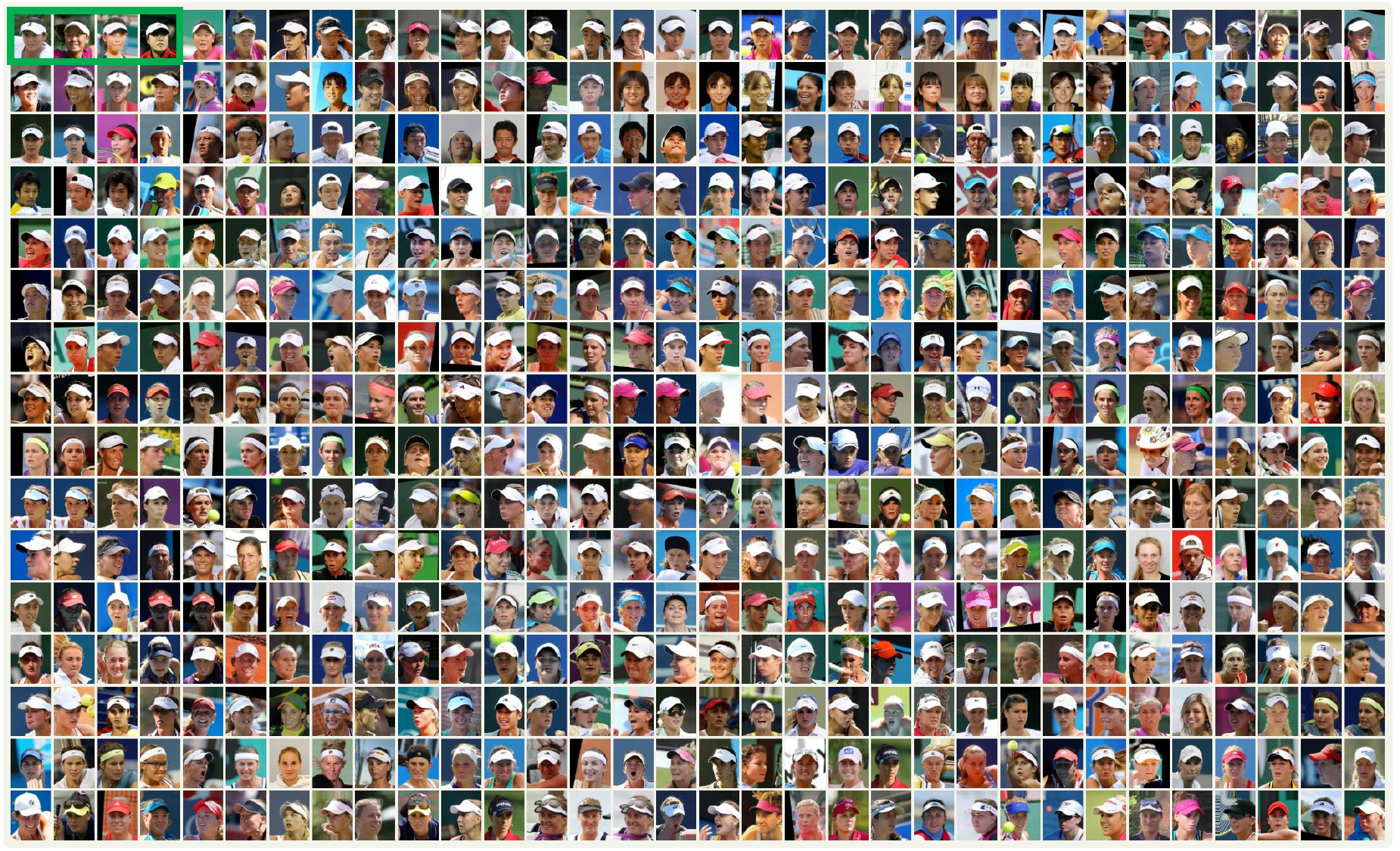}
    \caption{CelebA images of asian females with a tennis cap. Initial group in green.}
    \label{fig:g5}
\end{figure}

\section{Ablation and Sensitivity Analysis} \label{Appendix: Ablation Study}

\subsection{Sensitivity to Initialization and Projection Threshold $\tau$} \label{Appendix: Quantitative Ablation}

\begin{figure}[h]
    \centering
    \begin{subfigure}[b]{\textwidth}
        \includegraphics[width=\linewidth]{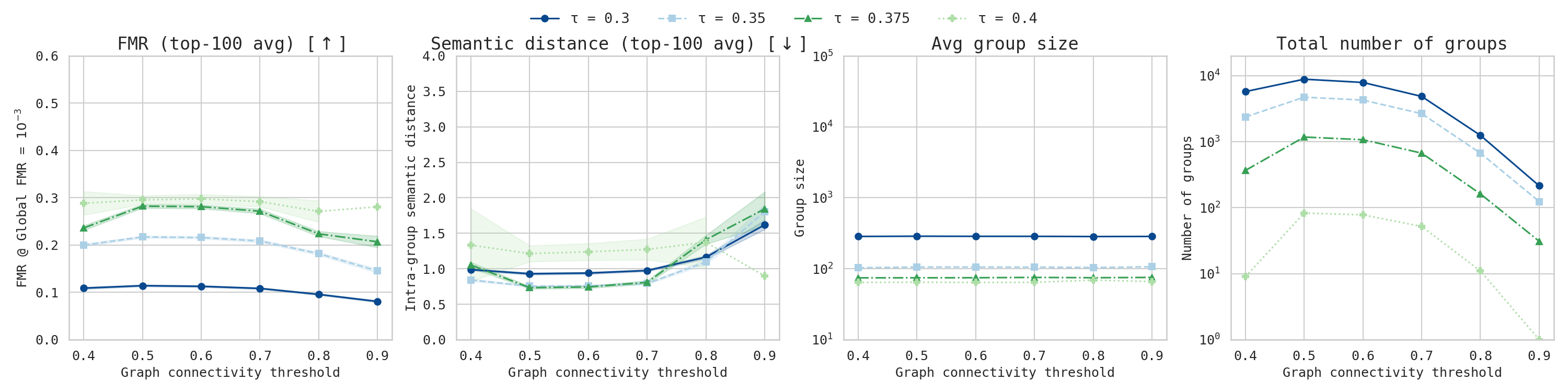}
        \caption{LatentAlign applied to \textbf{ArcFace}.}
    \end{subfigure}
    \begin{subfigure}[b]{\textwidth}
        \includegraphics[width=\linewidth]{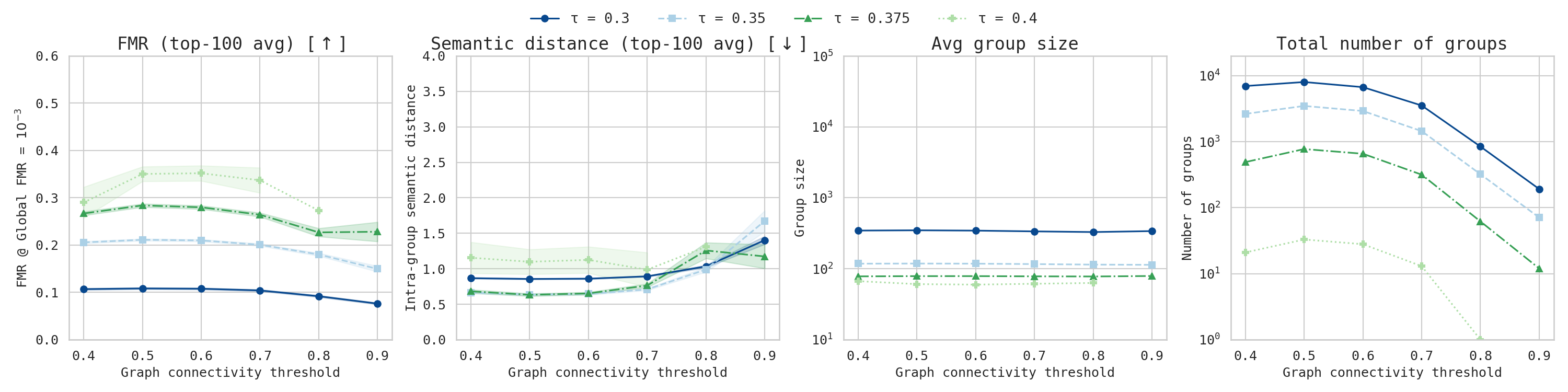}
        \caption{LatentAlign applied to \textbf{CosFace}.}
    \end{subfigure}
    \begin{subfigure}[b]{\textwidth}
        \includegraphics[width=\linewidth]{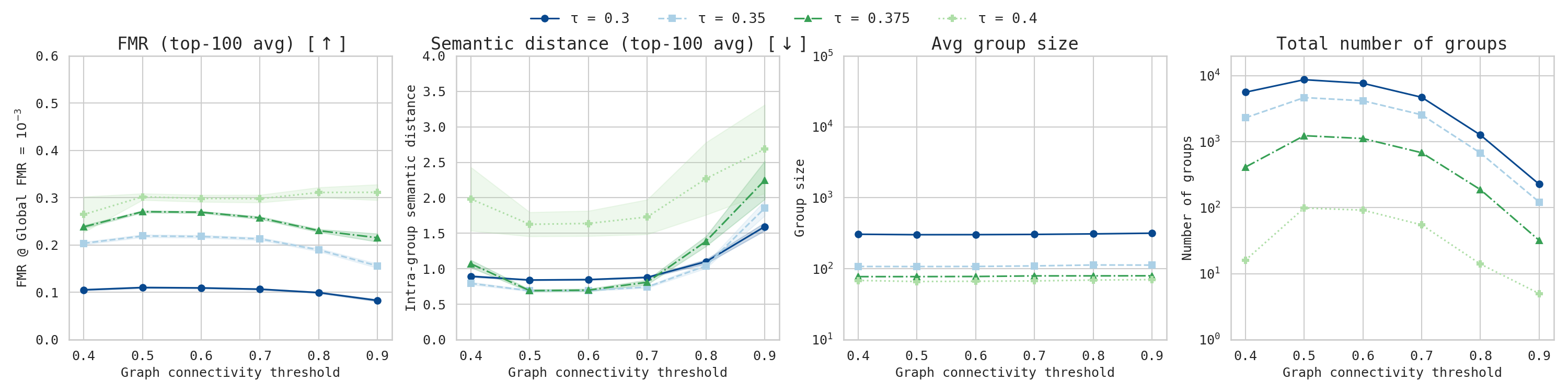}
        \caption{LatentAlign applied to \textbf{ElasticFace}.}
    \end{subfigure}
    \begin{subfigure}[b]{\textwidth}
        \includegraphics[width=\linewidth]{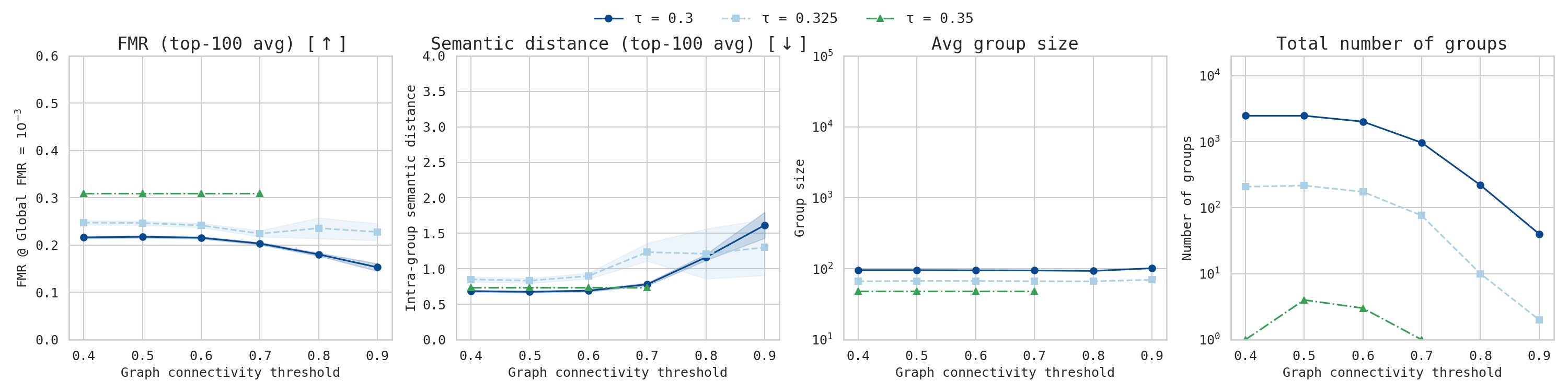}
        \caption{LatentAlign applied to \textbf{PartialFC}.}
    \end{subfigure}
    \caption{\textbf{LatentAlign parameter sensitivity.} Analysis of how the graph connectivity threshold and projection threshold $\tau$ affect FMR, semantic distance, and group formation. Results are shown for RFW embeddings obtained from \textbf{ArcFace}, \textbf{CosFace}, \textbf{ElasticFace} and \textbf{PartialFC} models.}
    \label{fig:LFA_ablation_1}
\end{figure}

\begin{figure}[h]
    \centering
    \begin{subfigure}[b]{\textwidth}
        \includegraphics[width=\linewidth]{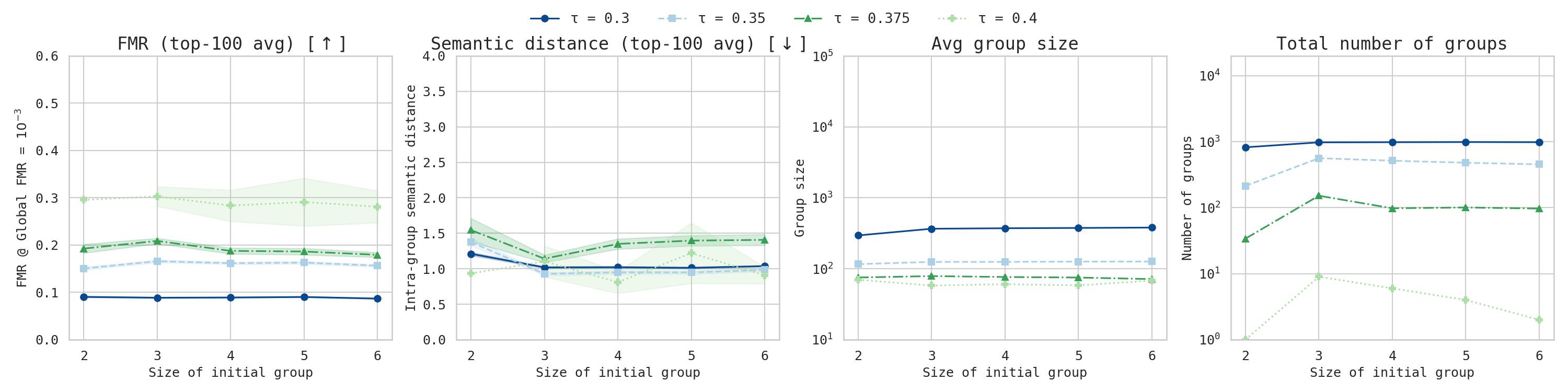}
        \caption{LatentAlign applied to \textbf{ArcFace}.}
    \end{subfigure}
    \begin{subfigure}[b]{\textwidth}
        \includegraphics[width=\linewidth]{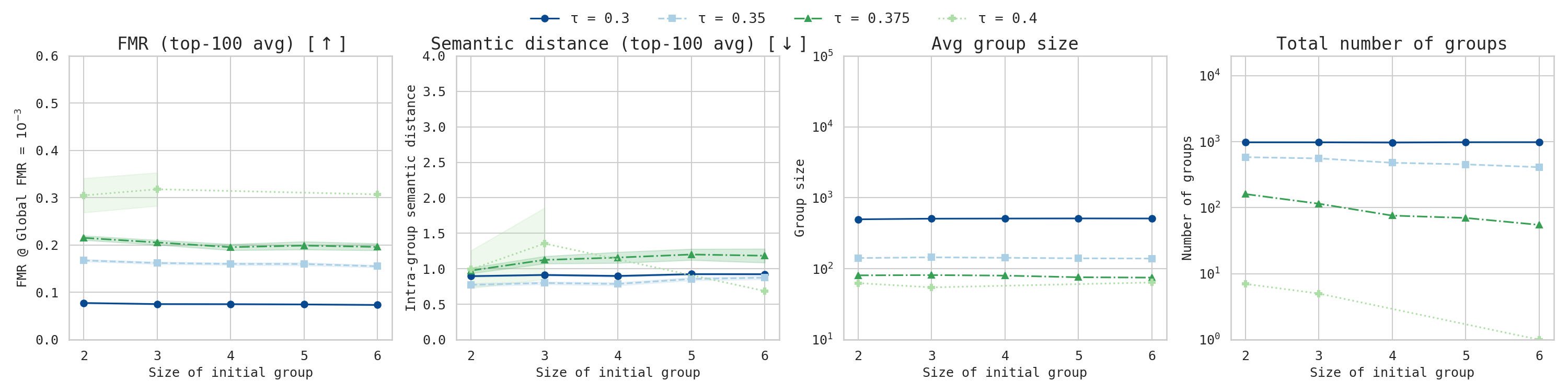}
        \caption{LatentAlign applied to \textbf{CosFace}.}
    \end{subfigure}
    \begin{subfigure}[b]{\textwidth}
        \includegraphics[width=\linewidth]{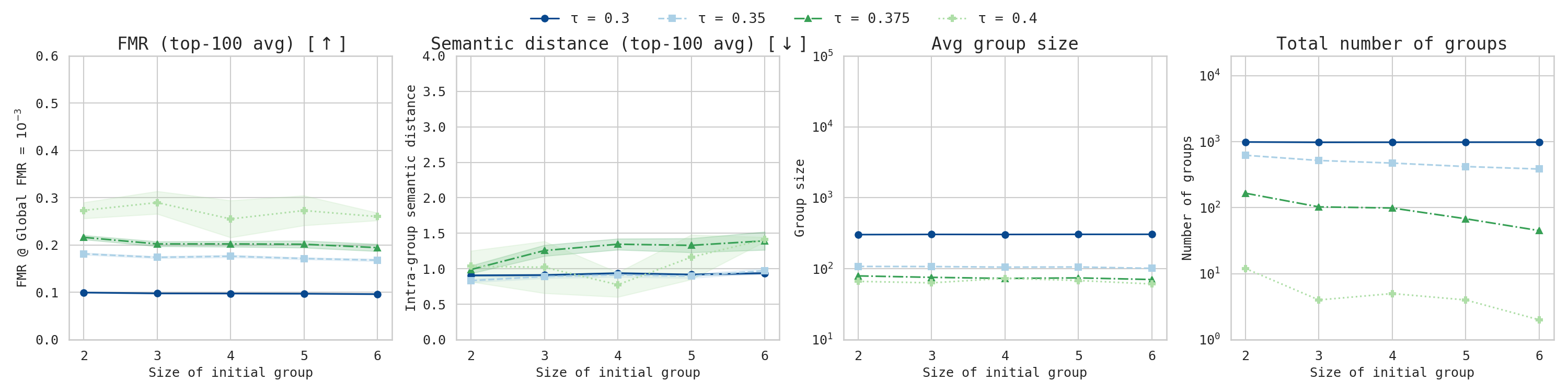}
        \caption{LatentAlign applied to \textbf{ElasticFace}.}
    \end{subfigure}
    \begin{subfigure}[b]{\textwidth}
        \includegraphics[width=\linewidth]{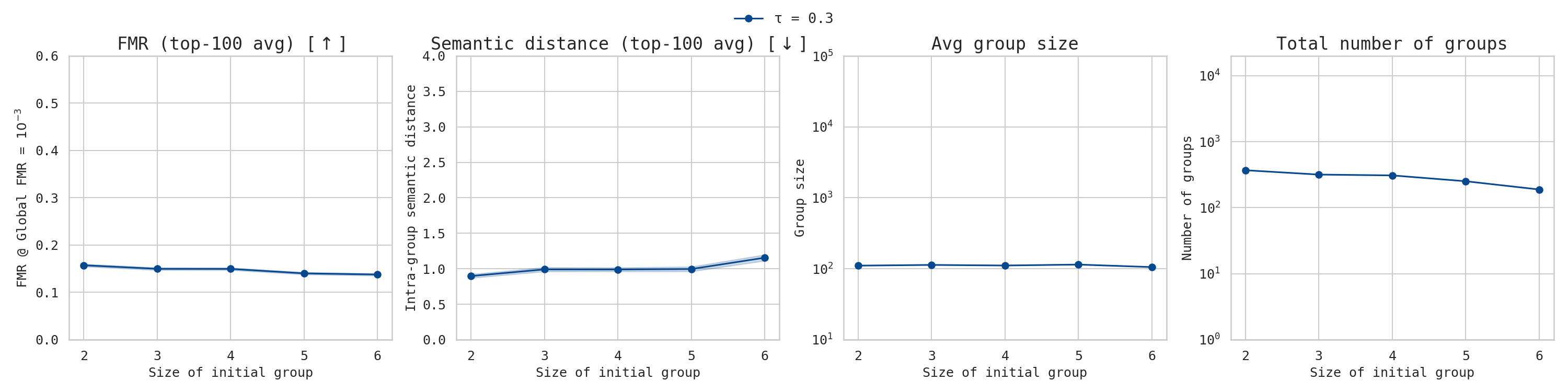}
        \caption{LatentAlign applied to \textbf{PartialFC}.}
    \end{subfigure}
    \caption{\textbf{LatentAlign sensitivity to random initialization.} Analysis of how the initial seed group size and projection threshold $\tau$ affect FMR, semantic distance, and group formation when using random seeds (without the similarity graph heuristic). Results are shown for RFW embeddings obtained from \textbf{ArcFace}, \textbf{CosFace}, \textbf{ElasticFace} and \textbf{PartialFC} models.}
    \label{fig:LFA_ablation_2}
\end{figure}

Figures \ref{fig:LFA_ablation_1} and \ref{fig:LFA_ablation_2} present an ablation study comparing our similarity graph initialization heuristic against random seeding. For the graph heuristic, we evaluate sensitivity across different graph connectivity thresholds, while for the random baseline, we vary the size of the initial seed group. Both approaches are tested across multiple projection thresholds $\tau$ using embeddings from the face recognition models.

To evaluate performance, we report four metrics: the average False Match Rate (FMR) of the top-100 most biased groups, the average intra-group semantic distance of the top-100 most coherent groups, the average group size, and the total number of discovered groups. FMR is computed at a global dataset threshold of $10^{-3}$.To ensure statistical validity, we restrict our evaluation to groups containing at least 20 distinct identities. Focusing explicitly on the top-100 groups aligns with our worst-case auditing framework, ensuring we evaluate the algorithms on their ability to expose the subpopulations that matter most.

The plots reveal a clear relationship between group size and FMR, whereas semantic distance remains relatively independent of group size. The projection threshold $\tau$ acts as a direct regulator here: a higher $\tau$ yields smaller, more tightly aligned groups that isolate the most severe intersectional vulnerabilities (the highest FMR). Conversely, a lower $\tau$ produces larger groups but results in a lower average FMR, as the bias becomes diluted across a broader set of attributes.

Notably, the results demonstrate that LatentAlign is highly robust to initialization. While the graph heuristic is computationally efficient and provides targeted starting points, random seeds ultimately converge to subpopulations with comparable top-100 FMR and semantic distance scores. This confirms that LatentAlign's core iterative directional alignment—rather than a fortunate initialization—is the primary driver for uncovering intersectional bias.

\clearpage

\subsection{Visualizing Semantic vs. Identity Alignment} \label{Appendix: Qualitative Ablation}

\begin{figure}[t]
    \centering
    \includegraphics[width=\linewidth]{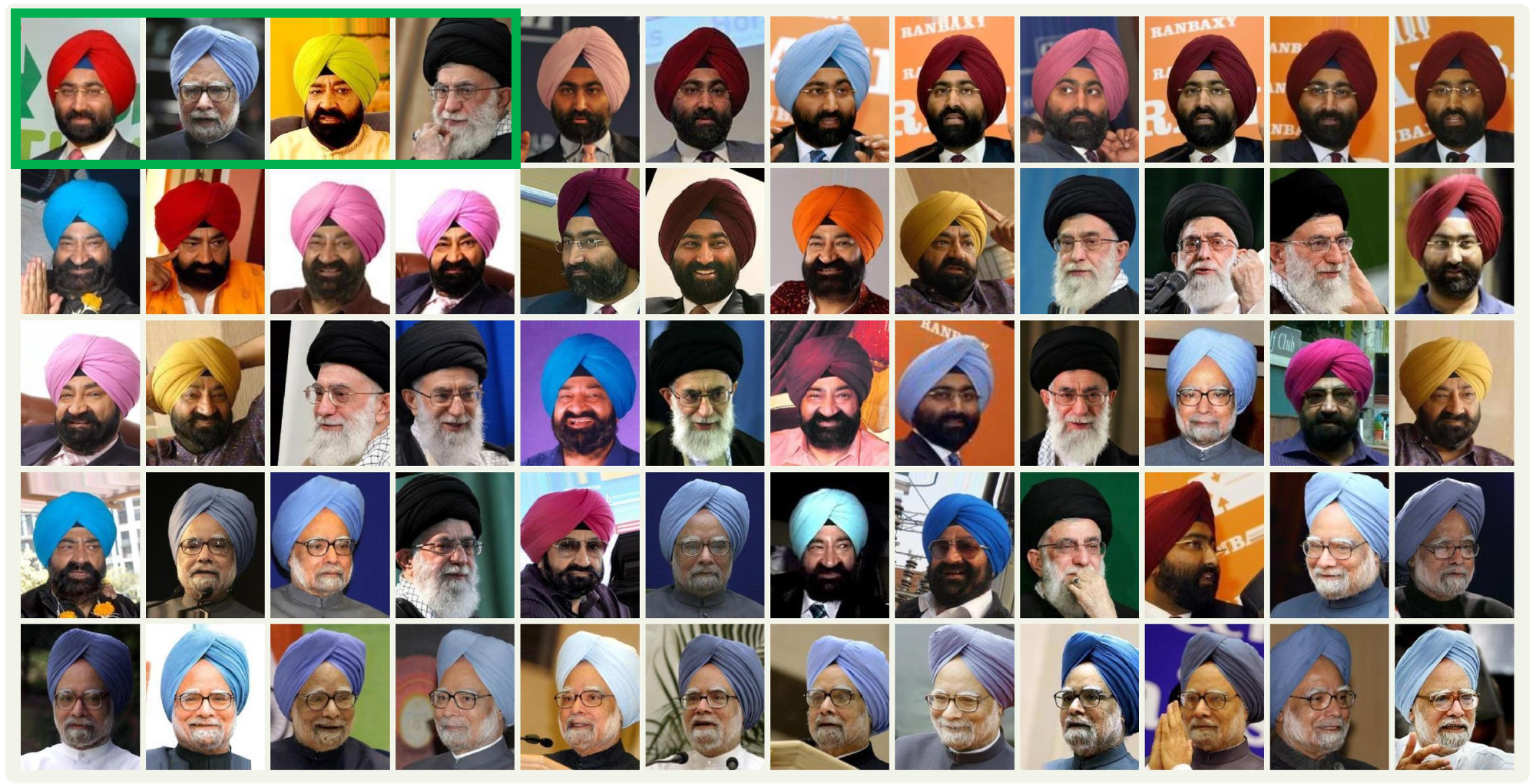}
    \caption{CelebA images of males with turban. Initial group in green.}
    \label{fig:g1}
\end{figure}

This section presents illustrative examples of high-quality groups formed by LatentAlign within the CelebA dataset. All groups were constructed using embeddings extracted with PartialFC and a projection threshold of $\tau = 0.35$. The examples highlight attribute-based groups that exhibit strong semantic coherence, driven by initializing the algorithm with images of different identities.

\begin{figure}[h]
    \centering
    \includegraphics[width=0.61\linewidth]{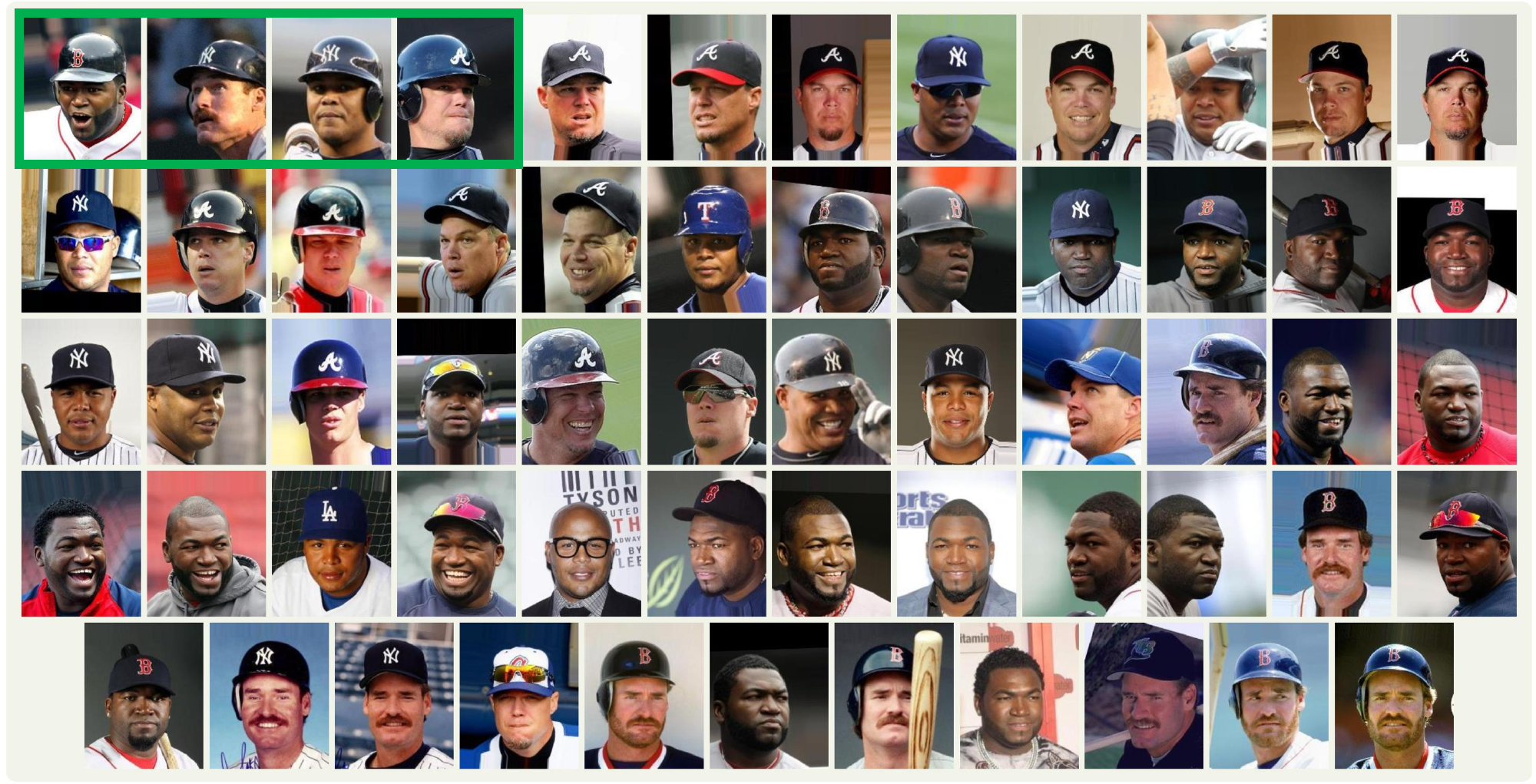}
    \caption{CelebA images of males with baseball helmet. Initial group in green.}
    \label{fig:g2}
\end{figure}

\begin{figure}[t]
    \centering
    \includegraphics[width=0.61\linewidth]{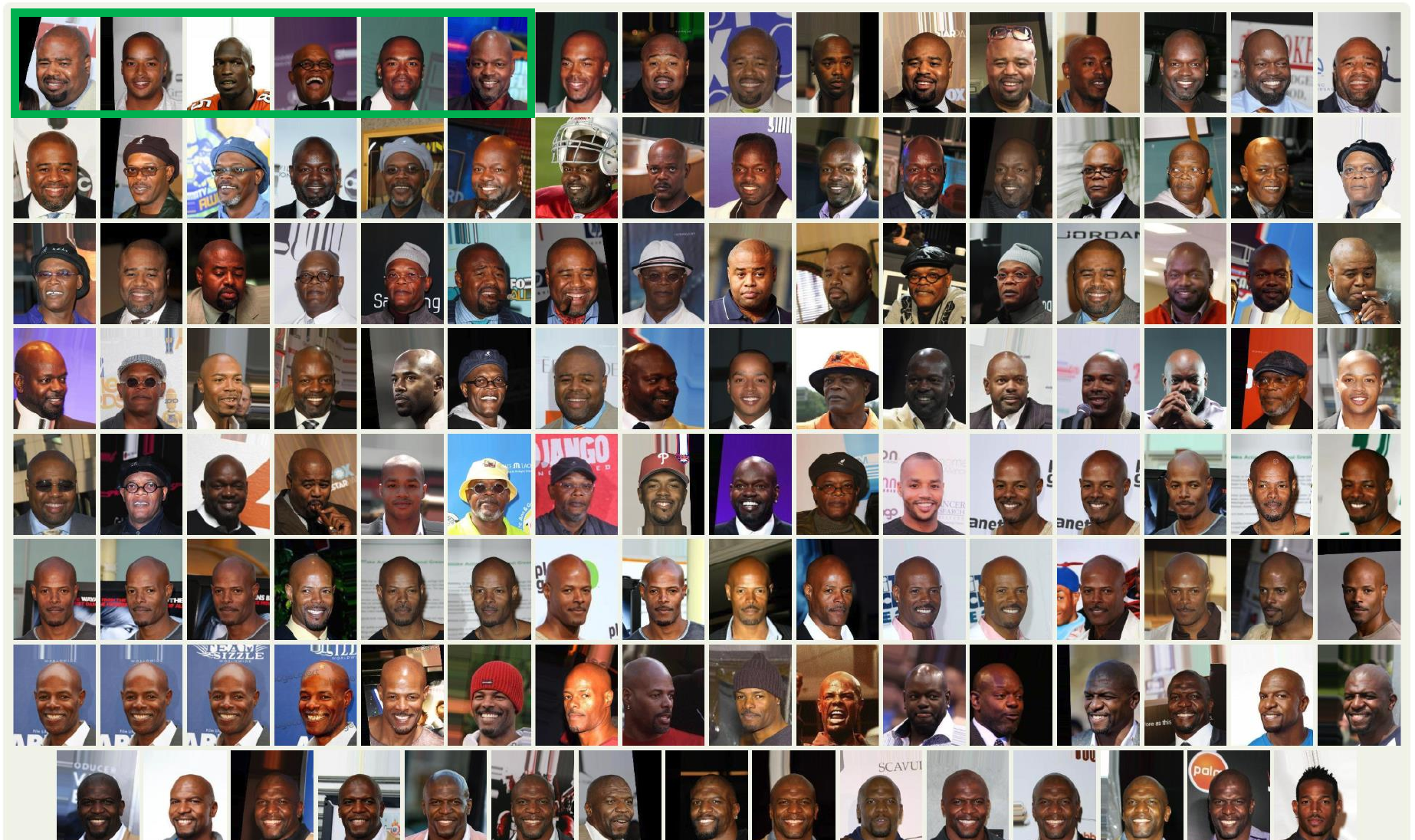}
    \caption{CelebA images of males with black skin and bald. Initial group in green.}
    \label{fig:g3}
\end{figure}

\begin{figure}[t]
    \centering
    \includegraphics[width=0.61\linewidth]{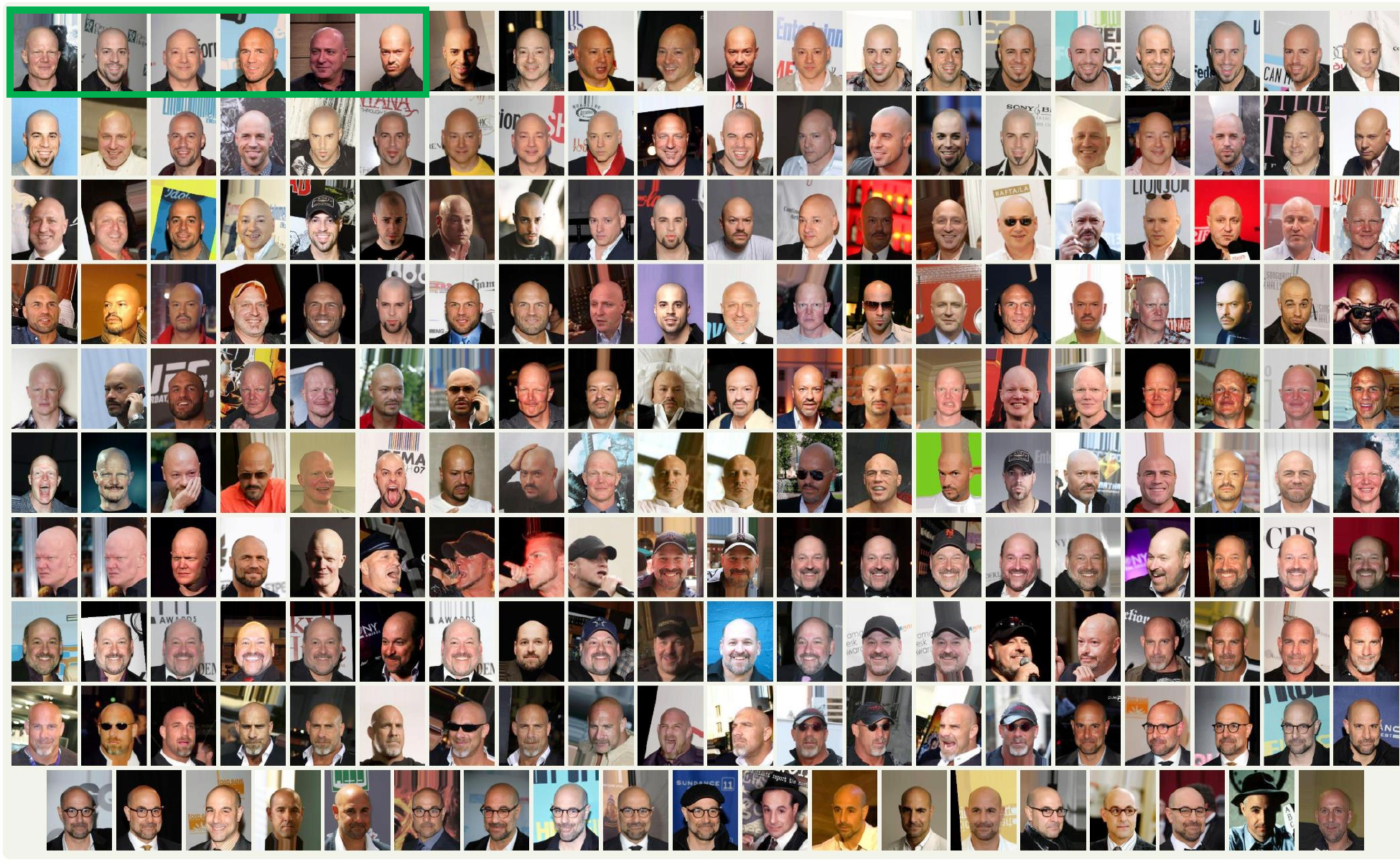}
    \caption{CelebA images of males with white skin and bald. Initial group in green.}
    \label{fig:g4}
    \vspace{1in}
\end{figure}

Figures \ref{fig:g1}--\ref{fig:g4} display groups representing distinct intersectional categories: males with turbans, males with baseball helmets, dark-skinned bald males, and light-skinned bald males. In each figure, the green box in the top-left corner indicates the initial seed group. Images are displayed in left-to-right, top-to-bottom sequence, reflecting the exact order in which the algorithm aligned and added them to the group.

We specifically focus this visual analysis on CelebA due to its high number of images per identity, which inherently increases the likelihood that grouping will be influenced by identity-specific features rather than shared semantic attributes. This is precisely the case when an initial group lacks identity diversity (e.g., consisting of images of the same person). In such cases, the iterative expansion tends to favor retrieving look-alike identities rather than aligning with a consistent, cross-identity attribute.

\begin{figure}[t]
    \centering
    \includegraphics[width=0.75\linewidth]{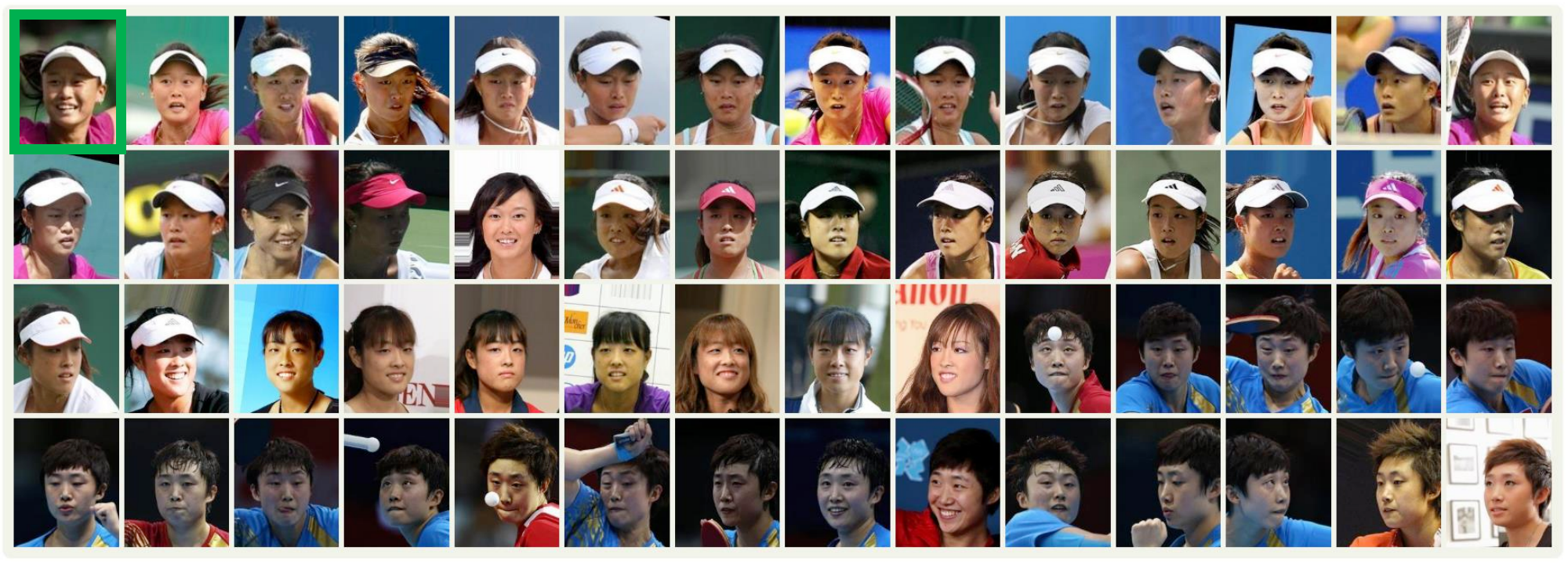}
    \caption{CelebA images of females. Single initial group in green.}
    \label{fig:fail1}
\end{figure}

\begin{figure}[t]
    \centering
    \includegraphics[width=0.75\linewidth]{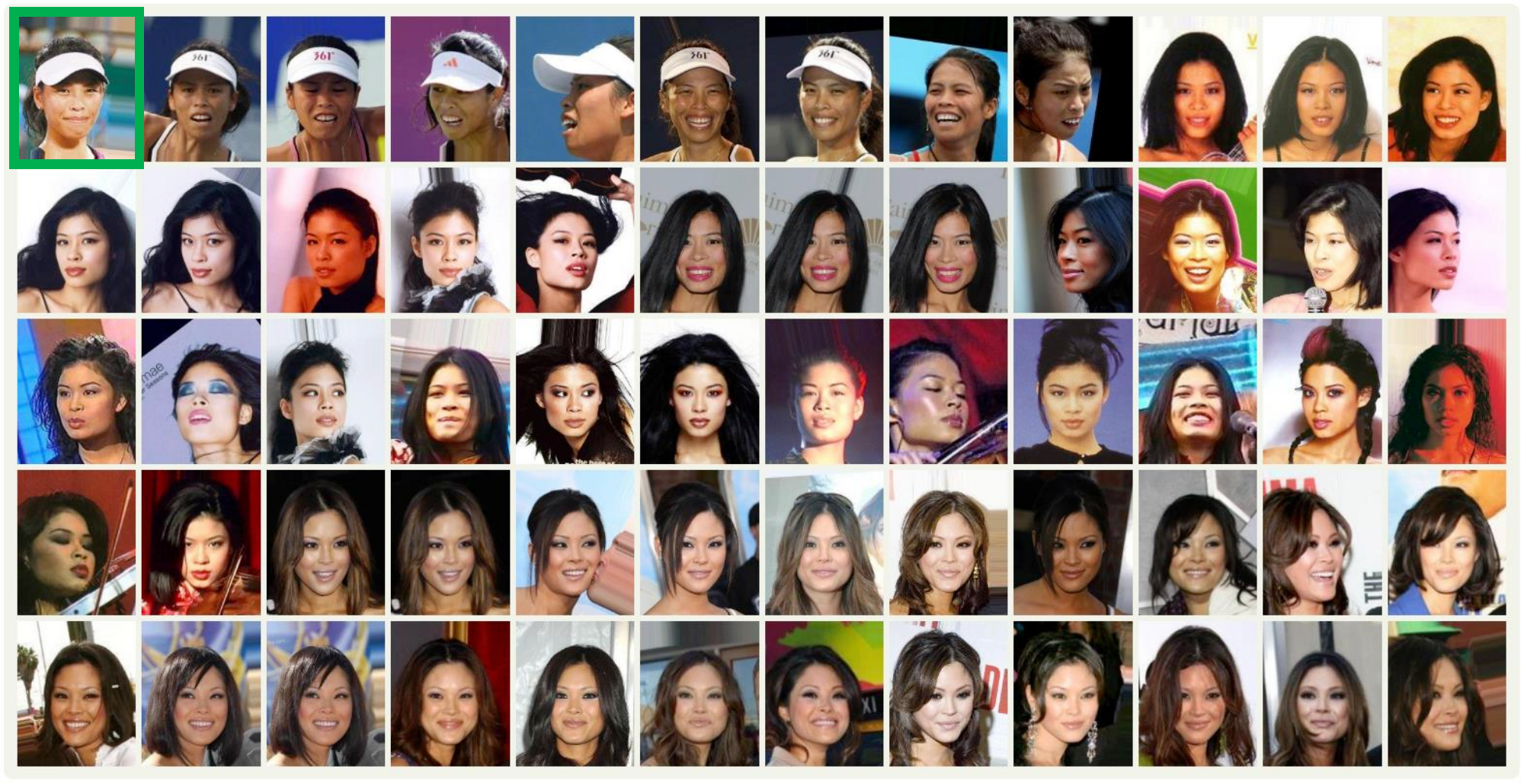}
    \caption{CelebA images of females. Single initial group in green.}
    \label{fig:fail2}
\end{figure}

\begin{figure}[t]
    \centering
    \includegraphics[width=0.75\linewidth]{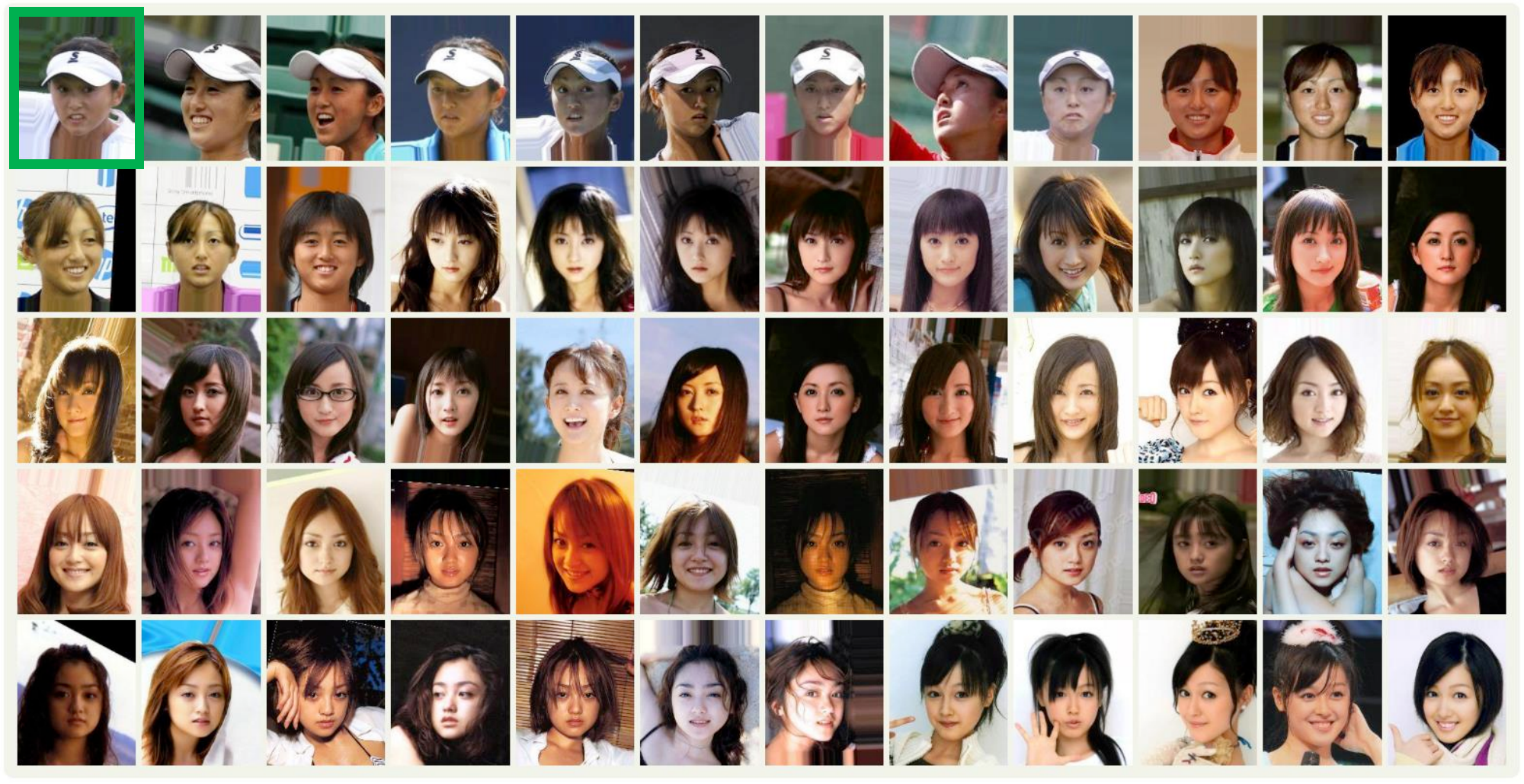}
    \caption{CelebA images of females. Single initial group in green.}
    \label{fig:fail3}
\end{figure}

To illustrate this failure mode, Figures \ref{fig:fail1}--\ref{fig:fail3} show examples where the initial group consisted of a single image. Here, the grouping is driven primarily by identity similarity. While not inherently undesirable, this leads to groups where the dominant shared feature is merely an identity-specific facial trait rather than a broader demographic or semantic variable.

Conversely, Figure \ref{fig:g5} illustrates the power of seed diversity. Initialized with four images of different Asian females wearing tennis caps, all from different identities, the resulting group consistently retrieves new individuals wearing (mostly white) tennis caps. This indicates that the latent direction successfully isolated and aligned with the intended visual attribute, bypassing identity.

Ultimately, these examples validate LatentAlign's ability to discover visually meaningful, attribute-driven groups when initialized with diverse faces. The discovered latent directions consistently reflect human-interpretable factors such as age, ancestry, attire, and hair. They also reinforce our claim that informed or exhaustive initial seed selection directly dictates grouping quality. Finally, while we observed some groups reflecting non-demographic semantics (e.g., a "black-and-white photograph" group combining stylistic and demographic factors in Figs. \ref{fig:subtraction} and \ref{fig:addition}), we did not observe groups dominated by pose or blur.



\clearpage

\begin{table}[t]
    \centering
    \caption{Comparison of intra-group attribute distances between $k$-means and LatentAlign at different levels of $N$ (average number of samples) in the CelebA database. Thresholds $\tau$ for LatentAlign and $k$ for $k$-means are shown in parentheses. Bold values indicate better performance between the two methods under comparable group sizes.}
    \label{tab:ablation-distance}
    \resizebox{\textwidth}{!}{%
    \begin{tabular}{lccccc}
      \textbf{Method} & \textbf{N (avg)} & \textbf{ArcFace} \small($\tau \mid$ k) & \textbf{CosFace} \small($\tau \mid$ k) & \textbf{ElasticFace} \small($\tau \mid$ k) & \textbf{PartialFC} \small($\tau \mid$ k)\\
      \cmidrule(lr){1-1}\cmidrule(lr){2-2}\cmidrule(lr){3-6}
      
      K-Means & $\sim$50  & 6.58 (4000) & 6.55 (4000) & \textbf{6.04} (4000) & \textbf{6.48} (4000) \\
      LatentAlign     & $\sim$50  & \textbf{6.09} (0.42) & \textbf{6.08} (0.42) & 6.21 (0.35) & 6.13 (0.40) \\
      
      \addlinespace
      
      K-Means & $\sim$100 & 7.65 (2000) & 7.61 (2000) & 7.55 (2000) & 7.50 (2000) \\
      LatentAlign     & $\sim$100 & \textbf{6.42} (0.40) & \textbf{6.41} (0.40) & \textbf{6.57} (0.33) & \textbf{6.28} (0.39) \\
      
      \addlinespace
      
      K-Means & $\sim$200 & 8.52 (1000) & 8.49 (1000) & 8.38 (1000) & 8.28 (1000) \\
      LatentAlign     & $\sim$200 & \textbf{6.87} (0.38) & \textbf{6.95} (0.38) & \textbf{7.03} (0.31) & \textbf{6.64} (0.37) \\
    \end{tabular}
    }
\end{table}

\section{Additional Baseline Comparisons}
\subsubsection{Size-Controlled Baseline Comparison on CelebA.}

While Figure \ref{fig:semantic_comparison} in the main text demonstrates the effect of varying cluster sizes on semantic distance for the RFW dataset, we extend this analysis here to the CelebA dataset. Specifically, we evaluate the effectiveness of LatentAlign against $k$-means clustering across all four face recognition models (ArcFace, CosFace, ElasticFace, and PartialFC).

A fair comparison requires both methods to produce subpopulations of similar average sizes, as larger groups inherently exhibit greater variation, and thus higher average intra-group attribute distance. Therefore, rather than selecting arbitrary hyperparameters, we systematically adjust $k$ (for $k$-means) and the projection threshold $\tau$ (for LatentAlign) to align the average cluster sizes. For instance, setting $k=2000$ on CelebA typically yields clusters of approximately 100 images.

In Table~\ref{tab:ablation-distance}, we report the average intra-group attribute distance (where lower is better) for both methods at three different cluster size settings: 50, 100, and 200 images per group. For each setting, the superior value for each model is bolded. The specific thresholds $\tau$ for LatentAlign and $k$ values for $k$-means used to achieve these comparable sizes are provided in parentheses.

\section{Implementation Details}\label{Appendix: Details}

All experiments were conducted on a single NVIDIA A100 GPU with 40GB of memory, an Intel Xeon IceLake-SP 8360Y CPU, and 512GB of RAM.

\textbf{Processing Times}
Labeling the RFW dataset using visual-language models required several days. Extracting face embeddings with face recognition models took several hours.

Our method was evaluated on two datasets:

RFW (40k images): Runtime ranged from approximately 10 minutes (with a low threshold) to a few hours (with a high threshold).

CelebA (200k images): Runtime ranged from a couple of hours (low threshold) to several hours (high threshold).

\end{document}